%% file: main.tex
\documentclass[unnumsec,webpdf,modern,large,square,numbers]{article}

\usepackage{arxiv}
\usepackage[utf8]{inputenc} 
\usepackage[T1]{fontenc}    
\usepackage{hyperref}       
\usepackage{url}            
\usepackage{booktabs}       
\usepackage{amsfonts}       
\usepackage{nicefrac}       
\usepackage{microtype}      
\usepackage{lipsum}		
\usepackage{graphicx}
\usepackage{natbib}
\usepackage{doi}
\usepackage{setspace}
\usepackage{algorithm}
\usepackage{amsmath}

\usepackage{inputenc}
\usepackage{subcaption}
\usepackage{graphicx,xcolor} 
\usepackage{multirow}
\usepackage{tabulary}
\usepackage[algo2e]{algorithm2e} 
\usepackage{algpseudocode}
\usepackage{enumitem}
\usepackage{svg}

\setitemize{labelindent=0.6em,labelsep=0.6cm,leftmargin=*}

\renewcommand{\headeright}{}
\renewcommand{\undertitle}{}
\begin{document}

\title{Benchmarking and Analyzing In-context Learning, Fine-tuning and Supervised Learning for Biomedical Knowledge Curation: a focused study on chemical entities of biological interest}


\author{
Emily Groves* \\
 Institute of Health Informatics\\
	University College London\\
	London, England \\
   \And
 Minhong Wang* \\
 Institute of Health Informatics\\
  University College London\\
	London, England \\
  \And
  Yusuf Abdulle \\
  Institute of Health Informatics\\
  University College London\\
	London, England \\
 \And
 Holger Kunz \\
  Institute of Health Informatics\\
  University College London\\
	London, England \\
 \And
 Jason Hoelscher-Obermaier \\
	iris.ai\\
	Oslo, Norway \\
 \And
 Ronin Wu \\
 iris.ai\\
 Oslo, Norway \\
 \And
 Honghan Wu \\
 Institute of Health Informatics\\
 University College London\\
 London, England \\
 \texttt{honghan.wu@ucl.ac.uk}
}

\date{}

\onecolumn 
\maketitle

\begin{abstract}
Automated knowledge curation for biomedical ontologies is key to ensure that they remain comprehensive, high-quality and up-to-date. In the era of foundational language models, this study aims to compare and analyze three natural language processing (NLP) paradigms for curation tasks: in-context learning (ICL), fine-tuning (FT) and supervised learning (ML).
Chemical Entities of Biological Interest (ChEBI) database was used as an exemplar ontology, on which three curation tasks were devised. GPT-4, GPT-3.5 and BioGPT were utilized for ICL using three prompting strategies. PubmedBERT was chosen for the FT paradigm. For ML, six embedding models were utilized for training Random Forest and Long-Short Term Memory models. To assess different paradigms' utilities in different data availability scenarios, five different setups were configured for assessing the effect on ML and FT performances.
The full datasets generated for curation tasks were task 1 (\#Triples 620,386), task 2 (611,430) and task 3 (617,381), with a 50:50 positive versus negative ratio. For ICL models, GPT-4 achieved best accuracy scores of 0.916, 0.766 and 0.874 for tasks 1-3 respectively. In a head-on-head comparison, ML (trained on around 260,000 triples) was more accurate than ICL in all tasks (accuracy differences: +.11, +.22 and +.17). Fine-tuned PubmedBERT performed similarly to best ML models in tasks 1 \& 2 (F1 differences: -.014 and +.002), but worse in task 3 (-.048). Simulation experiments showed both ML and FT models deteriorated in smaller and higher-imbalanced training data. When training data had 6,000 triples or fewer, GPT-4 was superior to ML/FT models in tasks 1 and 3. However, ICL never performed on par with the ML/FT in task 2.
When prompted properly, foundation models with ICL can be good assistants for knowledge curation, however, clearly not yet to a level making ML and FT paradigms obsolete. The latter two need good task-related training data to outperform ICL. Notably, in such situations, the ML paradigm only needs small pretrained embedding models and much less computation.
\end{abstract}

\keywords{Ontology enrichment, knowledge curation, In-context learning, Fine-tuning, Supervised learning, Foundataion models, LLM, GPT}

\pagebreak
\onehalfspacing
\section{Introduction}
Knowledge Graphs (KGs) ~\citep{hogan2021knowledge,pan2017exploiting} are a novel paradigm for integrating and representing semantically networked datasets or knowledge bases from highly heterogenous sources. They contain information about entities and their relationships. For instance, in directed KGs, knowledge may be represented as $(s, o, l)$ triples, where $s$ and $o$ respectively represent subject and object entities, and $l$ specifies the relationship, or link, associating subject and object entities.
As such, KGs are well-suited to the large and heterogeneous datasets common in the biomedical domain ~\citep{luo2016big}. Unsurprisingly, there is a large body of literature of utilising KGs for biomedical purposes. When integrated with patient-level data such as electronic health records (EHRs), KGs can facilitate automated diagnosis ~\citep{xu2019end,li2020graph} or generation of radiology reports ~\citep{zhang2020radiology}. Knowledge graph techniques have also shown great promise and utility in pharmaceutical and chemical studies, including novel adverse drug reaction predictions ~\citep{bean2017knowledge}, drug-drug interaction predictions ~\citep{lin2020kgnn} and drug repurposing ~\citep{zhang2021drug}.

KGs can suffer from sparsity and incompleteness~\cite{cheng2021communication}, and may also require updating periodically as new information becomes available. 
KG enrichment serves a valuable function in filling gaps to provide a more reliable, correct and complete representation of knowledge. In scenarios involving EHR data, the absence of enrichment can lead to a lack of contextual information, which is crucial for tasks like comparing diagnoses~\cite{carvalho2023knowledge}. 
Manual KG curation is, however, time-consuming, burdensome and impractical in settings where the pace of knowledge generation is high. Automated knowledge graph enrichment or refinement~\cite{paulheim2017knowledge} is a subfield in graph-based machine learning, focused on correctly integrating new entities into existing knowledge graphs and predicting novel relationships between entities. This can significantly enhance the efficiency of the curation process. For example, Kartsaklis and colleagues conducted a study to enrich KGs by employing multi-sense LSTMs, underlining the vital necessity to continue developing dynamic and comprehensive KGs~\cite{kartsaklis-etal-2018-mapping}. Furthermore, in a study focused on rare diseases, enrichment has been applied to identify sets of rare diseases that share aspects of pathophysiology and aetiology. This enrichment process aims to facilitate drug repurposing~\cite{sanjak2023clustering}.

Link prediction, among the most popular enrichment approaches, seeks to forecast new relationships between entities. When applied to biomedical knowledge graphs, link prediction becomes a versatile and powerful tool for hypothesis generation, particularly in domains such as drug discovery~\cite{abbas2021application}. Over the past decade, a plethora of machine learning link prediction algorithms have emerged. These often utilise readily accessible corpora as sources of complementary knowledge to populate and enhance KGs, often through linkage to resources like WordNet~\cite{rezayi2021edge}, Wikipedia, PubMed and Biomedical Literature, social media or common web crawl~\cite{kastrin_link_2016}. 
Due to the nature of these sources, it is not surprising to see Natural Language Processing (NLP) techniques~\cite{khurana2023natural} play an instrumental role in the creation and curation of KGs by automating effective information extraction at scale~\cite{wu2022survey}. An application of NLP in the knowledge creation process is found in KnowLife, a substantial knowledge repository in the fields of health and life sciences, automatically constructed from diverse web sources~\cite{ernst2015knowlife}. Similar efforts towards the construction of domain-specific knowledge graphs have prominently featured the employment of the Stanford CoreNLP tool to parse text into sentences, underscoring the expanding role of NLP in the realm of knowledge creation~\cite{yuan2020constructing}.

Among these applied NLP approaches, word embedding techniques~\cite{wang2019evaluating}, including the first generation of pretrained models such as Word2Vec~\cite{mikolovEfficientEstimationWord2013} and GloVe~\cite{pennington2014glove}, played a critical role in enhancing NLP tasks through \emph{semantic} representations of words. Technically, these techniques operate by representing words in textual content as numerical vectors, thereby attributing analogous semantic meanings to words with corresponding vector proximities. Such capacity was obtained by a so-called pretraining process, where large corpora are utilised in a self-supervision learning process. This has a clear utility in deriving \emph{new knowledge} for enriching KGs. For example, open domain embedding models have been shown to be useful in KG enrichment for recommendation systems~\cite{cao2021dekr}. Beyond general language understanding,  domain specific embeddings derived via pretraining on large domain corpora ~\cite{chen2019biosentvec,zhang2019biowordvec} are able to capture \emph{semantics}, showing very good understanding of domain knowledge, e.g., those~\cite{mao2020use,lu2021predicting} in the biomedical domain. 

Powerful, transformer-based Large Language Models (LLMs) have emerged in recent years~\cite{vaswani2017attention,wolf2020transformers}, significantly transforming Natural Language Processing. Via task-agnostic, self-supervised pre-training on vast corpora, these models learn lexical, syntactic, and semantic structures. One of the most representative ones is the Bidirectional Encoder Representations from Transformers (BERT)~\cite{devlin2018bert}. BERT was pre-trained on BooksCorpus (800M words) and English Wikipedia (2,500M words) through masked language modelling and next sentence prediction self-supervisions. The bidirectional transformer architecture enables the model to learn nuanced, contextualized word representations, yielding excellent results in natural language understanding and classification tasks.  With subsequent task-specific fine-tuning, BERT like models have been shown impressive performances in a range of downstream tasks including document classification, text completion, question answering and named entity recognition~\cite{sun_how_2019, yang_end--end_2019, lothritz_evaluating_2020}. 

Differing from the bidirectional architecture of BERT models, causal language models (CLMs) are another key type of LLM, whereby the model is given a sequence of words and is tasked with predicting the next word in the sequence. The Generative Pre-trained Transformer (GPT) series have demonstrated unparalleled capabilities in text generation~\cite{radford2018improving}. Further improved by the training technique such as ``in-context learning'', the GPT3 model demonstrates its strong performance simultaneously without fine-tuning on multiple NLP tasks, including translation, question-answering, as well as several tasks that require on-the-fly reasoning or domain adaptation, such as unscrambling words, using a novel word in a sentence, or performing 3-digit arithmetic~\cite{brown2020language}.

In late 2022, the public release of ChatGPT\cite{ChatGPT}, powered by GPT3.5/4 models surprised the general public with its ability to generate creative texts and wide range of capabilities, and is perceived by many as the dawn of artificial general intelligence.
The emergence of foundation models like GPT3.5/GPT4.0 and the open source LLMs (e.g., Llama2~\cite{touvron2023llama}, Mistral~\cite{jiang2023mistral}) has had an undeniable impact on NLP research. They are speeding up, or at least stimulating discussion on, paradigm shifts from supervised learning to fine-tuning to prompting/in-context learning~\cite{liu2023pre}, and from development of models specialised for a single function to versatile, general-purpose models which can be applied to a wide array of tasks. In this state of flux, for automated biomedical KG enrichment, it is sensible to ask:
\begin{itemize}
    \item How do these foundational LLMs perform in curating biomedical knowledge, including differences between models and effectiveness of various in-context learning strategies?
    \item Can smaller, domain-specific language models compete with large, open domain state-of-the-art LLMs?
    \item Are supervised learning approaches truly obsolete in such tasks?
    \item Is it still a useful research paradigm to conduct hypothesis-driven approaches on adapting small models (e.g., embedding models) for such tasks?
\end{itemize}

This study aims to conduct a series of experiments for answering these questions. The Chemical Entities of Biological Interest (ChEBI) database will be used for this focused study. ChEBI is one example of knowledge graphs in the interdisciplinary field between chemistry and biomedicine. It functions as both a database and an ontology, housing information about chemical entities of biological relevance, and contains a diverse array of curated data items~\cite{degtyarenko2007chebi,de2010chemical}. This resource finds extensive application across various domains, including drug target identification~\cite{liu2021ontological,rasheed2021identification} and gene studies~\cite{bettembourg2015optimal}. However, it is worth noting that in the case of ChEBI, the addition of new entities and connections is a manual process~\cite{moreno2015binche}, which translates to a substantial investment of time and resources.

\section{Materials and methods}
\begin{figure}[h]
\centering
\includegraphics[width=.83\linewidth]{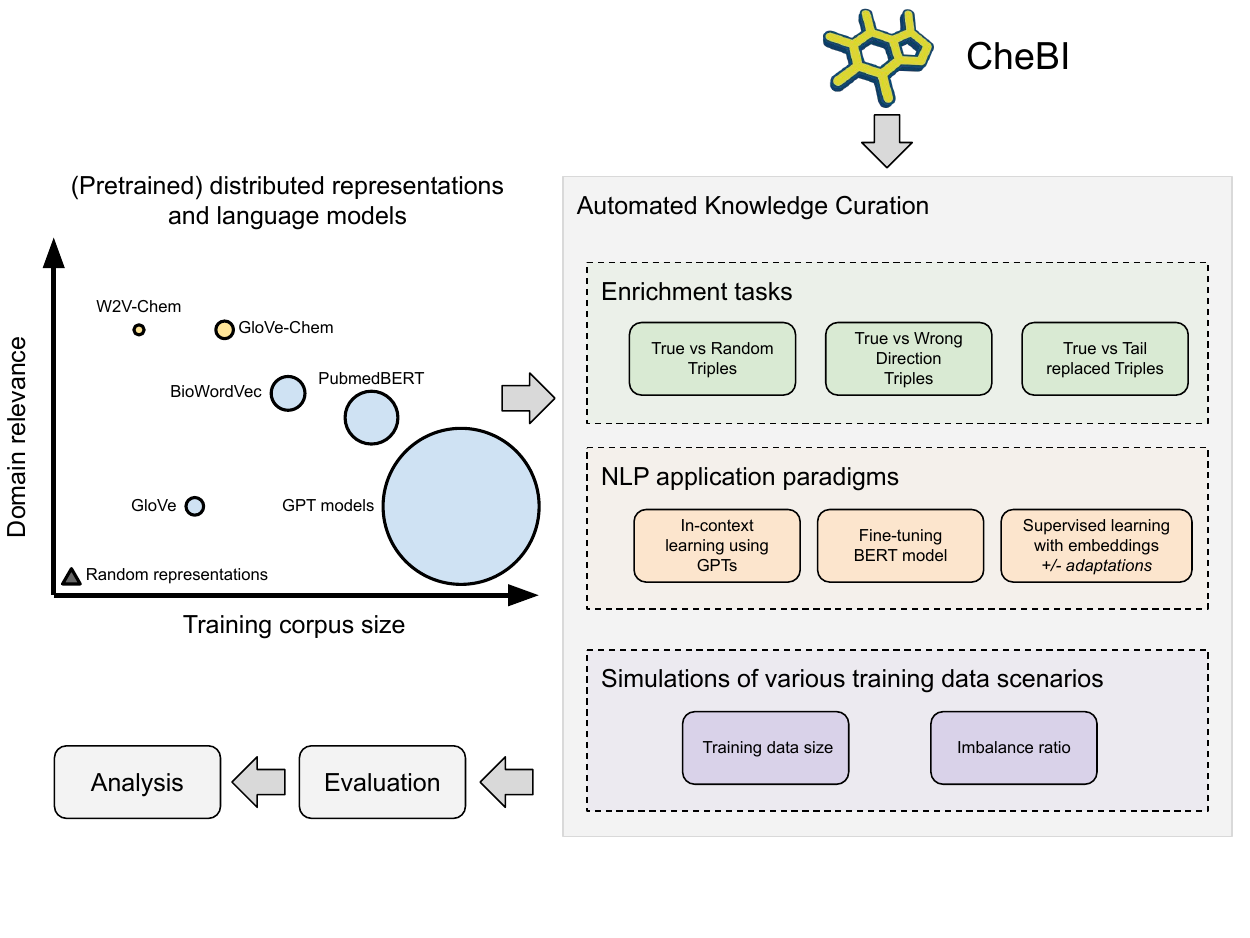}
\caption{Architecture of study design. Pretrained distributed representations and language models panel (top-left): X-axis is the size of corpus for pretraining the models. Y-axis is the relevance of the model pretraining corpus to the target domain. The size of the shapes denotes the model size. (Note: ratios are indicative only.) Three simulated knowledge curation tasks were devised, and different model adaptation techniques proposed. Different difficulty settings were adjusted via modification of training data size and imbalance ratio.}
\label{fig:fig1}
\end{figure}

A schematic representation of this work is provided in Figure \ref{fig:fig1}. The core (the box on the right of Figure~\ref{fig:fig1}) is a set of experiments to evaluate three paradigms of applying NLP in automated knowledge curation for the ChEBI KG: (1) in-context learning with pretrained large language models; (2) fine-tuning a pretrained BERT model with task-related training data; (3) supervised machine learning approaches using distributed representations, where hypothesis-driven adaptation was also applied. A diverse set of models (depicted in the two-dimensional space on the left) was utilized in this study, including large GPT models, a domain-specific BERT model, several pretrained embedding models and a random representation (i.e., embedding vectors were randomly generated). These models are plotted in the space based on their training corpus size (x-axis) and domain relevance (y-axis). The size of circles indicates model sizes (vocabulary size for embedding models and numbers of parameters for language models). \textit{NB: positions and sizes are indicative only.} Models which underwent further training on task related corpora are indicated in yellow. 

Three types of enrichment tasks were proposed to assess the models' abilities in detecting different forms of `erroneous' knowledge (the top dashed-line box on the right of Figure~\ref{fig:fig1}). We also simulated five scenarios where the size and imbalance (negative vs positive data labels) of the training data vary (the bottom dashed-line box in Figure~\ref{fig:fig1}). This is to assess how fine-tuning and supervised learning approaches perform in different situations, which will in turn give evidence on how to choose NLP paradigms, e.g., the settings where foundation models may be most useful. 

Details of these components in Figure~\ref{fig:fig1} are given in the rest of this section.

\subsection{ChEBI database}

The publicly available ChEBI knowledge graph\footnote{\url{https://www.ebi.ac.uk/chebi/downloadsForward.do}} was downloaded in  in February 2022 for comparing different approaches in knowledge graph enrichment tasks. We included data from all three sub-ontologies (see the appendix Table~\ref{tab.chebi}) and nine of the ten ChEBI relationship types (as listed in appendix Table~\ref{tab.rels}). For simplicity, the relationship `is conjugate acid of', which is the inverse relationship of `is conjugate base', was removed.

\subsection{Knowledge enrichment tasks} 
\label{tasks_summary}
The ChEBI ontology is denoted as $G=(V, T, L)$, where: 
\begin{itemize}
    \item $V$ is a set of nodes, representing all entities in ChEBI; 
    \item $T$ is a set of triples, where each triple $t = (s, o, l)$ consists of two nodes $s$ and $o$, called the subject and object of the triple, and a label $l$;
    \item $L$ is a set of labels, representing the possible relationships in ChEBI.
\end{itemize}

The enrichment task defined in this work is a simple binary classification$$f(t)=\begin{cases}
      1, & \text{if\ $t$ is a correct triple} \\
      0, & \text{otherwise}
    \end{cases}$$. A triple is called \emph{correct} if it denotes a true piece of knowledge.

Three binary classification tasks were devised in simulated knowledge enrichment tasks of identifying different types of erroneous triples:
\begin{itemize}
\item{\textbf{Task one (true vs random false triples)}: Positive triples comprised those extracted from the ChEBI database, i.e., $T_{pos} \subseteq T$. Negative triples were randomly generated as those without a directed link from a subject entity to an object entity, i.e., $T_{neg1}\subseteq\{(s,o,l)| s \in V, o \in V, (s,o,l) \notin T\}$.}
\item{\textbf{Task two (true vs wrong direction triples)}: This task assessed the degree to which models could distinguish true triples ($T_{pos}$) from flipped negative triples. Negative triples were generated by inversion of positive triples, $T_{neg2} \subseteq \{(o,s,l) | (s,o,l) \in T, (o,s,l) \notin T\}$. For example, for a positive triple \emph{(Androsta-4,9(11)-diene-3,17-dione, has\_role, androgen)}, the corresponding negative triple would be \emph{(androgen, has\_role, Androsta-4,9(11)-diene-3,17-dione)}.} 
\item{\textbf{Task three (true vs wrong object triples)} In the final and probably also most challenging task, models were asked to differentiate between true triples and their counterparts where only the object was replaced with a closely related entity, i.e., one of its sibling entities in ChEBI: $T_{neg3} \subseteq \{(s, o_2, l) | (s,o_1,l) \in T, (s, o_2, l) \notin T, p(o_1) \cap p(o_2)\neq \emptyset \}$, where $p(\cdot)$ is the function to retrieve parents of an entity. For example, for a positive triple \emph{(Androsta-4,9(11)-diene-3,17-dione, has\_role, androgen)}, a negative triple could be \emph{(Androsta-4,9(11)-diene-3,17-dione, has\_role, estrogen)}.}
\end{itemize}

\subsection{Pretrained language models and distributed representations} \label{word_embedding}
The following pretrained causal language models were included in this study for in-context learning experiments: 
\begin{itemize}
    \item BioGPT~\cite{luo2022biogpt}: a domain-specific generative Transformer language model pretrained on large-scale biomedical literature, comprising 15 million PubMed items (https://pubmed.ncbi.nlm.nih.gov), each with both title and abstract, retrieved before 2021.
    \item OpenAI's GPT model versions 3.5 and 4.0. Both models were built upon OpenAI's previous models~\cite{radford2019language,radford2018improvingb}. GPT3.5 was further improved based on InstructGPT~\cite{ouyang2022training} and Reinforcement Learning from Human Feedback (RLHF)~\cite{christiano2017deep}. The more recent and advanced GPT model, GPT4.0, represents one of the \emph{de factor} state-of-the-art foundation models. It has proven superior to its predecessors in many zero-short benchmark tests~\cite{espejel2023gpt}. Both models were accessed via OpenAI's API access point\footnote{\url{https://api.openai.com/v1/chat/completions}}. The GPT3.5 model used was gpt-3.5-turbo with model ID of \emph{gpt-3.5-turbo-0613}\footnote{\url{https://platform.openai.com/docs/models/gpt-3-5}} and the utilized GPT4.0 model name was gpt4 with an ID of \emph{gpt-4-0613}\footnote{\url{https://platform.openai.com/docs/models/gpt-4-and-gpt-4-turbo}}.
\end{itemize}

We used the PubmedBERT~\cite{gu2021domain} model in language model fine-tuning experiments. It was trained on PubMed corpus (accessed in 2020) with 14 million abstracts, 3.2 billion words and 21 GB texts. Notably, the model was trained on biomedical domain corpus from scratch.

For the third NLP paradigm experiments, supervised learning, the following (pretrained) embedding models were included. 
\begin{itemize}
\item{\textbf{GloVe~\cite{pennington2014glove}:} GloVE (Global Vectors for Word Representation) is an embedding model pretrained using an unsupervised learning algorithm on the Common Crawl corpus with a total of 840B tokens. This study used the \textit{glove.840B.300d} obtained from \url{https://nlp.stanford.edu/projects/glove/}, which has a 2.2M cased vocabulary and a vector dimension size of 300.}
\item{\textbf{W2V-Chem:} A word2vec~\cite{mikolov2013efficient} model was trained from scratch on 7,201 full papers from the chemical domain. These were sourced from PubMed using cross references associated with the ChEBI ontology. The titles, abstracts and full texts of these papers were used to train domain-specific word embeddings. Embeddings were initialized from random vectors.} 
\item{\textbf{GloVe-Chem:} We developed an embedding model by further adapting the GloVe embeddings for ChEBI enrichment. Specifically, the previously mentioned 7,201 PubMed articles were used to further train embeddings from the GloVe model. In contrast to W2V-Chem, the vocabulary was built by joining the texts from chemical domain papers and the vocabulary from GloVe. The input layer was initialised from Glove embeddings.}
\item{\textbf{Biowordvec~\cite{zhang2019biowordvec}:} An embedding model was developed for the biomedical domain using fastText~\cite{bojanowski2017enriching} trained on large biomedical corpus, as well as on information from the MeSH knowledge graph.}
\item{\textbf{PubmedBERT embeddings}: PubmedBERT was also used for deriving vector representations for triples. We summed up the last 4 hidden layers of the special token [CLS] for each component of a triple and used this as the entity representation.} 
\item{\textbf{Random embeddings:} We were also interested in evaluating word representations with no semantics by using a random embedding model. For a given ChEBI database entity, embeddings were generated via tokenization followed by assignment of a 300-dimension vector to each token. Vectors were randomly generated from uniform distribution between -1 and 1.}
\end{itemize}

\subsection{NLP paradigm 1: In-context learning with pretrained Large Language Models} \label{LLM_prompting}

Relative performances of GPT-3.5 Turbo, GPT-4 and BioGPT in the three binary classification tasks were evaluated using few-shot prompting. 
Models were provided with three positive and three negative example triples, and prompted to classify a seventh. Triples used in prompts were selected randomly from training data, eliminating any duplicates. Approximately equal numbers of positive and negative triples were queried for classification. 

For each task, models were provided with 100 distinct prompts, with each repeated five times. We experimented with three prompt formulations: 
(i) a base prompt (Table \ref{tab:Prompt formulations}); (ii) a second variant in which we added an additional sentence \emph{‘If you do not know the answer, state `I don't know’}, aiming to reduce hallucinations, and hence improve the utilities and performances; and (iii) a variant in which positive and negative example triples were presented in a random  order. The latter was done in response to an observed tendency for the BioGPT model to disproportionately classify triples as negative, having hypothesized that this might result from the order in which triples were presented (three positive examples, three negative examples, query triple). 

\begin{table}[h]
\begin{tabular}{@{} p{16.5cm} @{}}
\hline
\textbf{Basic prompt template} \\ \hline
\end{tabular}
\begin{verbatim}
"""  
Your task is to classify triples as True or False.  
<triple>: {positive_example_1} 
<classification>: True  
<triple>: {positive_example_2} 
<classification>: True  
<triple>: {positive_example_3} 
<classification>: True  
<triple>: {negative_example_1} 
<classification>: False  
<triple>: {negative_example_2} 
<classification>: False  
<triple>: {negative_example_3} 
<classification>: False  
<triple>: {prompt_triple}   
"""
\end{verbatim}
\begin{tabular}{@{} p{16.5cm} @{}}
\\ \hline
\end{tabular}
\caption{\label{tab:Prompt formulations}The template used for prompting LLMs (Variant \#1). In generating prompts, curly bracket-enclosed contents were replaced with corresponding triples derived from actual data.}
\end{table}

\subsection{NLP paradigm 2: fine-tuning BERT model for knowledge curation}
The PubmedBERT model was fine-tuned to conduct each of the three classification tasks. Triples were converted into sequences of words by concatenating the labels of subject, relationship and object with a special separator token \emph{$<$SEP$>$}. The sequence was then tokenized using the PubmedBERT tokenizer, and the output was fed into the transformer layers. The final layer of the model is a fine-tuning layer specific to the document classification task. This layer takes the output of the encoder layer and applies a feed-forward neural network to produce a vector representation of the document. This vector representation is then passed through a softmax layer to produce a probability distribution over the possible document classes (true or false in our scenario).

\subsection{NLP paradigm 3: supervised learning using embedding models}

\begin{algorithm}[]
\caption{supervised learning for knowledge curation using embeddings}\label{alg:supervised_learning}
\SetKwInOut{Input}{input}\SetKwInOut{Output}{output}
\Input {$X$: the training data - a list of triples; \\
$y$: the labels of the training data; \\
$emb$: an embedding model; \\
$sep\_token$: a special token as separator; \\
$Tkn$: a tokenizer; \\
$M$: the supervised learning algorithm; \\
$model\_type$: the type of the $M$ \\
} 

\Output {fitted model $m$}
$X_v \gets []$ \\
$v_{sep} \gets Emb(sep\_token)$\\
\For{$t$ in $X$}{
$(s, o, l) \gets t$\\
\If {$model\_type$ is RNN}{  \Comment{For RNN like algorithms, generate a sequence of vectors }
$(w_{s1}, ..., w_{si}) \gets Tkn(s)$ \\
$(w_{l1}, ..., w_{lj}) \gets Tkn(l)$ \\
$(w_{o1}, ..., w_{om}) \gets Tkn(o)$ \\
$vect \gets [emb(w_{s1}), ..., emb(w_{sj}), v_{sep}, emb(w_{l1}), ..., emb(w_{lj}),  v_{sep}, emb(w_{o1}), ..., emb(w_{oj})]$ \\
}
\Else{ \Comment{for others (e.g., Random Forest), average embeddings for subject, relationship and object, then concatenate them}
$vect \gets concatenate(\frac{\sum_{w \in Tkn(s)}{emb(w)}}{|Tkn(s)|}, \frac{\sum_{w \in Tkn(l)}{emb(w)}}{|Tkn(l)|}, \frac{\sum_{w \in Tkn(o)}{emb(w)}}{|Tkn(o)|})$\; 
}
$X_v.append(vect)$\; \\
}
$m \gets M.fit(X_v,y)$\; \\
return $m$\;
\end{algorithm}

For supervised learning approaches, we follow the same process as formalized in Algorithm~\ref{alg:supervised_learning} for using different embedding models and different learning algorithms. Essentially, triples were converted to vector representations, which were then fed into the chosen machine learning (ML) algorithm for model fitting. The vector representations were generated depending on the chosen algorithm type:
\begin{itemize}
    \item If the ML algorithm is a recurrent neural network (RNN) or its variants (e.g., LSTM - Long Short Term Memory network), the representation of a triple will be a sequence of vectors generated by (1) tokenizing each component of the triple; (2) converting each token into a vector using the chosen embedding model; (3) merging three vector sequences by using a special separator token, indicating the boundaries of the components.
    \item For other algorithms, triples will be converted into one vector by (1) tokenizing each component of the triple; (2) averaging vectors of each component; (3) combining the representations of each of the three components by concatenation.
\end{itemize}

LSTM and Random Forest ML algorithms were chosen in our implementations, representing RNN and non-sequential archetypes. Two types of tokenizers were used; for the PubmedBERT embedding model, the PubmedBERT tokenizer was used. For all other embedding models, we used the the NLTK~\cite{bird2006nltk} library. Specifically, its RegexpTokenizer~\footnote{\url{https://www.nltk.org/api/nltk.tokenize.RegexpTokenizer.html}} was used with hand-crafted regular expression patterns for tokenizing special chemical entity names. Random vectors were used for out of vocabulary situations. Hyperparameter optimization was applied using a 5-fold cross validation on training data, optimized for F1-scores. Appendix Table A7 details the grid search space for parameters.


\subsection{Hypothesis driven adaptation: token analysis and selection} \label{sec:adaptations}
Surprisingly, in our initial results using the random forest algorithm, we observed that random embeddings outperformed all semantic embeddings (see the \textbf{No adaptation} column of Table~\ref{tab:ml_task1}). Hence, we carried out further analysis to investigate this counter-intuitive finding.
Analysis of feature importance generated by random forest models revealed that, among models trained on non-contextual semantic embeddings, little weighting was given to head (i.e., subject of a triple) entities. In contrast, the model trained using random embeddings paid greater attention to head entities (see the figures on the left of the supplementary Figure~\ref{fig:feature_importance_task1_allTokens}). Based on this observation, we examined the high-frequency tokens in head (subject) and tail (object) entities and discovered that head entities contained many similar, short tokens that were likely to be non-meaningful for decision making (e.g., 1, 2, 6r, 2s etc. see Appendix Table~\ref{tab:res_top_frequent_tokens} for the 20 most frequent tokens). For random embeddings, they are easily separable because their representations are randomly generated. However, semantic representations of these tokens would be very close to each other in the vector space when using semantic-rich embedding models. 

\SetKwComment{Comment}{/* }{ */}
\begin{algorithm}[]
\caption{Algorithm for embedding-specific identification of less semantically meaningful tokens}\label{alg:token_selection}
\SetKwInOut{Input}{input}\SetKwInOut{Output}{output}
\Input {$token\_freq$: the dictionary of tokens and their frequency among positive triple head and tail entities\; \\
$emb$: the dictionary of word embedding\; \\
$E$: the list of all unique entities in all positive triples\; \\
$N$: the number of randomly selected entities, set as 5000\; \\
$I$: the number of iterations, set as 10\; \\
} 

\Output {stop words $S$}
$S \gets []$\; \\
$T \gets top\_frequent(token\_freq)$\; \Comment*[r]{Select top 25\% frequent tokens} 
$Clusters \gets get\_clusters\_DBScan(T,emb)$ \Comment*[r]{Get clusters of the tokens based on the tokens' embeddings}
\For{$i$ in $1...I$}{
\For{$cluster$ in $Clusters$}{
$E_N \gets random\_sample(E, N)$ \Comment*[r]{Randomly select N entities}
$M1 \gets get\_embedding(E_N, emb)$\;  \\
$M2 \gets get\_embedding(E_N, emb, cluster)$ \Comment*[r]{Get the centroid of the embeddings of the entities with or without the tokens in the cluster}
$D1[t].add(get\_distance\_variance(M1))$\; \\
$D2[t].add(get\_distance\_variance(M2))$ \Comment*[r]{Get the distance variance based on the pairwise distance in M1 and M2 separately}
}
$i \mathrel{+}= 1$
}
\For{$cluster$ in $Clusters$}{
$p \gets t\_test(D1[cluster], D2[cluster])$ \Comment*[r]{Two sample t-test based on 10 runs results}
\If{$p \leq 0.05$}{
$S.add(tokens\_in(cluster))$\;
}
}
return $S$

\end{algorithm}

We therefore hypothesized that high-frequency tokens with similar semantic meanings would bring entity representations closer in the vector space, making it more challenging for ML models to distinguish the relationship between them, i.e., less informative in classification tasks. Based on this hypothesis, we tested two algorithms for mitigating such an effect:
\begin{itemize}
 \item{\textbf{Naive adaptation:} Which removes tokens on the basis of their length. We only included tokens of 3 or more characters in generation of entity representations. Where all tokens in the entity name were shorter than 3 letters, we included all tokens.}
 \item{\textbf{Task-oriented adaptation:} Which uses a token clustering and t-test based approach to token selection, described in Algorithm~\ref{alg:token_selection}. This algorithm is able to adapt token selection based on embedding models. Briefly, the top 25\% most frequently seen tokens in positive triple head and tail entities were identified and grouped into clusters using a clustering algorithm (DBScan~\cite{schubert2017dbscan} in our case). Across 10 iterations, 5000 unique head and tail entities were randomly sampled. Entity representations were calculated as the centroid of token embeddings both with and without tokens from each identified cluster ($M1$ and $M2$, respectively). Based on our hypothesis, if tokens in a given cluster resulted in a significant difference between M1 and M2, then the tokens in this cluster were identified as stop words.}
\end{itemize}

When doing adaptation, one of the two token selection algorithms will be applied to filter the list of tokens after the tokenization step in the supervised learning algorithm (Algorithm~\ref{alg:supervised_learning}). The effects of these adaptations will be compared with adaptation-free ML models and other paradigms of NLP for the enrichment tasks.

\subsection{Effects of imbalanced data and variations in training data size}
It is well understood that ML model performances are adversely affected by training data imbalance and/or scarcity. We therefore sought to explore and compare model performances (paradigms 2 \& 3: fine-tuning and supervised learning) under these sub-optimal conditions. Such an investigation would also provide evidence on when the paradigm 1 is most useful, since pretrained LLMs have been trained on large corpora and thus have minimal dependency on task-related training data. 

Using reduced datasets (\(\sim \)10\% of the full training and test datasets), we generated varying train-test split ratios via random selection of successively smaller subsets of training data (9:1, 8:1, 7.1, 6:1, 5:1, 4:1, 3:1, 2:1, 1:1, 0.5:1 for training:testing splits). Effects of imbalanced data were determined by altering the ratio of positive versus negative triples present in training data (1:1, 0.75:1, 0.5:1, 0.25:1, 0.125:1 positive:negative).

\section{Results}


\subsection{The statistics of the ChEBI ontology}
As of February 2022, ChEBI contained 147,461 entities. Chemical Entities represent the majority (145,869),  followed by 1,550 Role Entities and 42 Subatomic Particles. There are total 318,438 triples, with a highly imbalanced distribution of relationships; the 3 most common types make up $>90\%$ of all triples: 230,241 (72.3\%) \emph{is\_a}, 42,095(13.2\%) \emph{has\_role} and 18,204 (5.7\%) \emph{has\_functional\_parent} (see Appendix Table \ref{tab:summary_relationships_freqs} for details). 

A total of 47,701 unique tokens were derived from these triples using the NLTK tokenization process described in the method section. Appendix Table \ref{tab:embeddings_basic} shows vocabulary sizes, numbers of embedding dimensions and proportions out-of-vocabulary tokens for each of the word embedding models included in this study. As might be expected, domain-specific word embeddings trained from large corpora of scientific literature (e.g. Biowordvec) have less out-of-vocabulary tokens compared to generic word embeddings (e.g GloVe).

\subsection{Data preprocessing and experiment setups} \label{sect:preprocessing}
\input{result_data_preprocessing}

\input{tables}
\input{figures}

\subsection{Results of supervised learning paradigm} 
\label{exp:ml}
\input{result_ml}

\subsection{Results of fine-tuning PubmedBERT} \label{exp:finetuning}
We fine-tuned the PubmedBERT model for three document classification tasks, and utilised a Cross-Entropy loss function. This model ran for 3 epochs and found that there was negligible differences in performance when running between 3 and 4 epochs, and thus optimised for performance. The learning rate for this model was set to $1\times10^{-4}$ and employed the Adam optimiser. The fine-tuning datasets for three tasks and results of fine-tuned PubmedBERT are summarised in Table~\ref{tab:ret_finetuned}. Overall, performances are on par with Random Forest model trained on PubmedBERT embeddings, and frequently rank among the best approaches tested, although not consistently so. In particular, the fine-tuned PubmedBERT model for task 3 was about 5\% worse than the best ML based model (Random forest using GloVe-Chem with naive adaptation).

\subsection{Results of In-context learning paradigm: prompting three GPT models} \label{exp:prompt}
\input{result_llms}

\subsection{Comparisons of three NLP paradigms on enrichment tasks} \label{exp:comparison_overview}
\input{results_comparison_3paradigms}

\section{Discussion}
We set out to implement automated approaches for curating knowledge for the ChEBI ontology in a transformative time for NLP research and development, where the medical informatics communities are in the heat of discussions on paradigm shift of applying NLP technologies in biomedicine. In this study, we followed three typical NLP paradigms, implemented a number of models using eight pretrained models, and conducted a series of experiments on three knowledge enrichment tasks. These have generated comprehensive sets of results and revealed some insightful findings.

In the head-on-head comparisons, the state-of-the-art foundation models, GPT-4, didn't perform well in our three tasks. The best NLP paradigm seems to be supervised learning methods using domain/task related pretrained distributed representations. Fine-tuning pretrained language models also performed strongly, particularly in task 2. However, such interpretations might only translate to situations where there is sufficient training data for the task on hand, as the ML models and fine-tuned models were trained or fine-tuned on plenty of data (at the scale of hundreds of thousands of triples). Further experiments simulating five data availability scenarios revealed more detailed and practical insights. For tasks 1 and 3, GPT-4 was clearly superior when the training data contained no more than 6,000 triples with an imbalance around 1:8 (positive:negative) or higher. However, GPT models seemed particularly poor in task 2 (i.e., classifying wrong relationship directions), where their in-context learning capacities never surpassed other NLP paradigms in all five scenarios tested.

Among the three GPT models, the domain specific BioGPT was not as good as generic counterparts. Recall scores were particularly poor. It also tends to give irrelevant answers even when prompted not to do so. This may have been due to the significant differences of training corpus size and number of parameters, and also the fact that it was not further improved via reinforcement learning from human feedback. GPT-3.5 and GPT-4 also showed very consistent results reflected by their Kappa scores. Prompting these two models not to make a classification when unsure led to considerably high performances (F1 scores: 0.79-0.91) for those triples where a definitive classification was made. Combined, these observations indicate that state-of-the-art foundation models could be very promising tools for knowledge curation, albeit leaving a 5-11\% of the data unclassified.

Fine-tuning pretrained language models was shown to be an effective approach for enrichment tasks 1 and 2. When there was abundant data, its performances were among the best. Fine-tuned models performed much stronger compared to supervised ML models when there was only 9\% of the full dataset for training, i.e., at the scale of 55,000. They were also shown to have the greatest consistency as training data availability was decreased. However, the fine-tuning approach, at least with the PubmedBERT model used in this work, seemed to bear some shortcomings regarding task 3 in our simulation experiments. Although its performance was initially strong (using 9\% of the full dataset for training), its performances deteriorated much faster with a F1-score of 0.47 in the fifth scenario (training data: 3,087 triples and 1:9 imbalance ratio). The reason behind this observation is worthy of further investigation, and may lead to some interesting findings.

Our results showed that supervised learning using distributed representations was certainly still a valid NLP paradigm for knowledge curation tasks. When abundant training data was available, even ML using random embeddings could achieve very good performances, which were superior to in-context learning using GPT models. Task-specific pretrained embedding models (trained on ChEBI related articles) were clearly very useful to such curation tasks, achieving the best performances in the various setups explored. In particular, W2V-Chem embeddings - only trained on around 7,000 PubMed articles - achieved surprisingly good performances. This demonstrates the effectiveness of a simple model (word2vec) with a small task-related corpus in downstream tasks. 

Hypothesis driven adaptation of embedding models was shown to be very effective; adaptations for filtering and selecting tokens helped to generate the best models in all three tasks. They also showed the ability to increase models' consistency when having less training data in our simulation experiments, where ML models with adaptations were among the top models. Between the two adaptation strategies, the naive approach was consistently better than the more sophisticated token selection algorithm in full-dataset experiments. The reason could be related to the dataset uniqueness and some of the hyperparameters in the task-oriented adaptations. However, the simulation experiments showed task-oriented adaptation was better in all tasks. Further analysis on this observation would assist in the development of better token selection algorithms, which in turn would further improve supervised learning based approaches for knowledge curation. Nevertheless, it was positive to see evidence that hypothesis driven approaches combined supervised learning are still effective and valid in the LLM era.

A key limitation of this work was that only a single ontology/KG was utilized, potentially leading to questions on the generalizability of these findings. To mitigate this, our study introduced three different types of curation tasks, and assessed model performances in five different data availability scenarios. Combined, these generated 15 different scenarios, representing a comprehensive exploration of the effectiveness of these approaches in a diverse range of settings. Nevertheless, future work using diverse datasets would produce more conclusive findings across different application domains. The other limitation was the potential reproducibility issue caused by the use of OpenAI's GPT models via their API access. It is well known that these models are continually undergoing revision and improvement. For example, our initial GPT-3.5 experiments conducted in July 2023 yielded significantly poorer results than the latest run on the same model in November 2023. Future work should evaluate the use of open source GPT models like Meta's Llama2~\cite{touvron2023llama}. 

\section{Conclusion}
This work evaluated three NLP research paradigms in the context of knowledge curation for enriching biomedical ontologies: in-context learning using large language models, fine-tuning pretrained BERT-based language models and `traditional' supervised learning using pretrained embeddings. We used the ChEBI database as an exemplar ontology. Six different pretrained embedding models were used, two of which were trained specifically for this study using a dedicated chemistry corpus. Extensive experiments were conducted via three different curation tasks, with five scenarios of data availability and imbalance for training/fine-tuning. Based on the in-depth analysis of these findings, we believe we now have evidence to support answers to the four research questions, which we set out to address.

\begin{itemize}
    \item How do these foundational LLMs perform in curating biomedical knowledge, including differences between models and effectiveness of various in-context learning strategies? \\
    \textit{Answer: In-context learning using the state-of-the-art LLMs did yield the best performance. However, they do have their utilities if proper prompting strategies are used. Even in those situations, there are still cases that these models could not give confident results. Also, in some situations they are particularly poor like our task 2, classifying triples with wrong relationship directions.}
    
    \item Can smaller, domain-specific language models compete with large, open domain state-of-the-art LLMs? \\
    \textit{Answer: short answer, yes. Longer answer, it depends. Our study shows fine-tuning a domain specific language model, PubmedBERT, outperformed GPT-4 when there is abundant training data - more than 24,000 triples in all tasks.}
    
    \item Are supervised learning approaches truly obsolete in such tasks? \\ 
    \textit{Answer: definitely not. Again, this depends on the availability of your training data. When the training dataset is at a scale of 50,000, supervised learning models even only trained on random vectors showed better performances. In general, domain/task-related embedding models achieved better or comparable performances to those of prompted LLMs with around 24,000 triples for training. It is notable that such models are much smaller and cheaper to run.}
    
    \item Is it still a useful research paradigm to conduct hypothesis-driven approaches on adapting small models (e.g., embedding models) for such tasks? 
    \\ \textit{Answer: yes. We showed hypothesis-driven adaptation approaches in using embedding models significantly improved performances in all tasks and all data availability scenarios.}
\end{itemize}

\section{CONFLICT OF INTEREST STATEMENT}
No competing interest is declared.

\section{ETHICS APPROVAL}
The study involved no human participants.

\section{AUTHOR CONTRIBUTIONS STATEMENT}
H.W., M.W., E.G., R.W. conceived the study and experiments; M.W., E.G., Y.A. collected, preprocessed the data and developed the NLP models; E.G., M.W., H.W. analysed the results; H.W., M.W., E.G., Y.A. drafted the manuscript; all authors reviewed and approved the final manuscript.

\section{DATA AVAILABILITY STATEMENT}
The ChEBI is publicly available from EBI's official website. The preprocessed data can be reproduced using our code and the version generated in this work will also be shared on reasonable request to the corresponding author. The code of this study will be available on Github after peer review.

\section{FUNDINGS}
This work was supported by UK's Medical Research Council (MR/S004149/1, MR/X030075/1); National Institute for Health Research (NIHR202639); British Council (UCL-NMU-SEU International Collaboration On Artificial Intelligence In Medicine: Tackling Challenges Of Low Generalisability And Health Inequality); Iris.AI - The AI Chemist (Research Council of Norway); UCL Global Engagement Fund 2022/2023; HW’s role in this research was partially funded by the Legal \& General Group (research grant to establish the independent Advanced Care Research Centre at University of Edinburgh). The funders had no role in conduct of the study, interpretation, or the decision to submit for publication. The views expressed are those of the authors and not necessarily those of Legal \& General. 

\bibliographystyle{plain}
\bibliography{main}

\section{Appendix}
\input{appendix}

\end{document}

%% file: result_data_preprocessing.tex

Here, we describe the process of generating training/test datasets for all three tasks (Section~\ref{tasks_summary} gives formalized definition of this.). As all experiments are binary classification tasks, the purpose of the dataset generation process was to create positive and negative triples.
\begin{itemize}
    \item Task 1. Positive triples used all triples from ChEBI except those with relationship \emph{is conjugate base of}. This led to a total of 310,193 positive triples. The same number of negative triples were created by randomly generating triples using ChEBI entities. 
    \item Task 2. Positive triples used all positive ones from task 1, except those with a relationship of \emph{is tautomer of} because it is a symmetric relationship. This yielded 305,715 positive triples. For each positive triple, we simply swapped the subject and object to create a negative triple. This effectively generated the same number of negative triples.
    \item Task 3. The same set of positive triples from Task 1 was used. For each positive triple, a negative triple was created by changing the object with one of its siblings. If no sibling entity existed, or the process resulted in formation of a true triple, no negative triple was generated. A total of 307,188 negative triples were created in this manner.
\end{itemize}

The \textbf{Triples} column of Table~\ref{tab:datasets} shows the detailed numbers of the populated datasets. Derived from this same base, the data used for three NLP paradigms were as follows.

\begin{itemize}
    \item \textbf{Machine learning on embedding models}: a stratified random split with a 9:1 ratio was used to populate training and test datasets for all three tasks. Table~\ref{tab:datasets}'s \textbf{Training set} and \textbf{Test set} columns show the detailed numbers for each tasks.
    \item \textbf{Fine-tuning the PubmedBERT model} a stratified random split with a 8:1:1 ratio was used to populate training, validation and test sets, detailed numbers in Table~\ref{tab:ret_finetuned}.    
    \item \textbf{In-context learning on GPT models}: for each task, 50 positive triples were drawn at random from the positive subset of the full dataset. Similarly, another 50 were drawn from the negative subset. All triples chosen are of the relationship type \emph{is\_a}. We selected only shorter (possibly less complex) triples by including those with less than 60 tokens. In order to evaluate the consistency of LLMs' classifications, prompts were each delivered five times before calculation of the Fleiss' Kappa statistic.
    \item \textbf{Comparison experiments between three paradigms} for each task, a 50 positive and 50 negative triples were randomly drawn from the positive and negative subsets of the full \textbf{Test set}. Note that we did not put constraints on the relationship types in this comparison.
\end{itemize}

%% file: tables.tex
\begin{table}[]
\centering
\begin{tabular}{l|ll||ll|ll||l}
\hline
       & \multicolumn{2}{c||}{Triples}                 & \multicolumn{2}{c|}{Training set}            & \multicolumn{2}{c||}{Test set}                & \multirow{2}{*}{Total} \\ \cline{1-7}
       & \multicolumn{1}{l|}{\#positive} & \#negative & \multicolumn{1}{l|}{\#positive} & \#negative & \multicolumn{1}{l|}{\#positive} & \#negative &                        \\ \hline
Task 1 & \multicolumn{1}{l|}{310,193}    & 310,193    & \multicolumn{1}{l|}{279,178}    & 279,177    & \multicolumn{1}{l|}{31,015}     & 31,016     & 620,386                \\ \hline
Task 2 & \multicolumn{1}{l|}{305,715}    & 305,715    & \multicolumn{1}{l|}{275,146}    & 275,146    & \multicolumn{1}{l|}{30,569}     & 30,569     & 611,430                \\ \hline
Task 3 & \multicolumn{1}{l|}{310,193}    & 307,188    & \multicolumn{1}{l|}{279,178}    & 276,469    & \multicolumn{1}{l|}{31,015}     & 30,719     & 617,381                \\ \hline
\end{tabular}
\caption{Statistics of generated datasets for three tasks. Training and test sets shown are for the supervised learning paradigm, which was based on a split of 9:1 ratio.}
\label{tab:datasets}
\label{tab:ml_task1}
\end{table}

\begin{table}[h]
\centering
\begin{subtable}[h]{\textwidth}
\centering
\begin{tabular}{l|ccc|ccc|ccc} \hline
 \multirow{2}{*}{Embeddings}   & \multicolumn{3}{c|}{No adaptation}                                                         & \multicolumn{3}{c|}{Naive adaptation}                                                      & \multicolumn{3}{c}{Task-oriented adaptation}                                              \\ \cline{2-10}
           & \multicolumn{1}{l}{Precision} & \multicolumn{1}{l}{Recall} & \multicolumn{1}{l|}{F1-Score} & \multicolumn{1}{l}{Precision} & \multicolumn{1}{l}{Recall} & \multicolumn{1}{l|}{F1-Score} & \multicolumn{1}{l}{Precision} & \multicolumn{1}{l}{Recall} & \multicolumn{1}{l}{F1-Score} \\ \hline
Random     & \textbf{0.9561}               & \textbf{0.9557}            & \textbf{0.9559}              & 0.9576                        & 0.9573                     & 0.9574                       & -                             & -                          & -                            \\
GloVe      & 0.9096                        & 0.9067                     & 0.9081                       & 0.954                         & 0.9536                     & 0.9538                       & 0.9606                        & 0.9604                     & 0.9605                       \\
W2V-Chem   & 0.9160                        & 0.9157                     & 0.9158                       & {\underline{\textbf{0.9691}}}         & {\underline{ \textbf{0.969}}}       & {\underline{\textbf{0.9690}}}        & 0.959                         & 0.9589                     & 0.9589                       \\
GloVe-Chem & 0.9192                        & 0.9186                     & 0.9189                       & 0.9683                        & 0.9683                     & 0.9683                       & 0.9198                        & 0.9194                     & 0.9196                       \\
BioWordVec & 0.9300                        & 0.9298                     & 0.9299                       & 0.9676                        & 0.9675                     & 0.9675                       & \textbf{0.9674}               & \textbf{0.9673}            & \textbf{0.9673}              \\
PubmedBERT & 0.9356                        & 0.9353                     & 0.9354                       & -                             & -                          & -                            & -                             & -                          & -          \\ \hline                 
\end{tabular}
\caption{Random Forest models' performances on Task 1 with different adaptation methods. Bold text indicates the best performance in each setup. Bold and underlined ones indicate those best performances overall in the task.}
\label{tab:ml_task1}
\end{subtable}

\begin{subtable}[h]{\textwidth}
\centering
\begin{tabular}{l|ccc||ccc} \hline
\multirow{3}{*}{Embeddings} & \multicolumn{3}{c||}{Task 2}                                                                & \multicolumn{3}{c}{Task 3}                                                                \\ \cline{2-7}
                  & \multicolumn{3}{c||}{Naive adaptation}                                                      & \multicolumn{3}{c}{Naive adaptation}                                                      \\ \cline{2-7}
                  & \multicolumn{1}{l}{Precision} & \multicolumn{1}{l}{Recall} & \multicolumn{1}{l||}{F1-Score} & \multicolumn{1}{l}{Precision} & \multicolumn{1}{l}{Recall} & \multicolumn{1}{l}{F1-Score} \\ \hline
Random            & 0.9581                        & 0.9581                     & 0.9581                       & 0.9042                        & 0.9042                     & 0.9042                       \\
GloVe             & 0.9573                        & 0.9573                     & 0.9573                       & 0.9073                        & 0.9073                     & 0.9073                       \\
W2V-Chem          & 0.9596                        & 0.9596                     & 0.9596                       & 0.9122                        & 0.9122                     & 0.9122                       \\
GloVe-Chem        & 0.9586                        & 0.9586                     & 0.9586                       & {\underline{\textbf{0.9126}}}         & {\underline{\textbf{0.9125}}}      & {\underline{\textbf{0.9125}}}        \\
BioWordVec        & 0.9605                        & 0.9605                     & 0.9605                       & 0.9062                        & 0.9061                     & 0.9061                       \\
PubmedBERT        & {\underline{\textbf{0.9822}}}         & {\underline{\textbf{0.9822}}}      & {\underline{\textbf{0.9822}}}        & 0.906                         & 0.906                      & 0.9060             \\ \hline         
\end{tabular}
\caption{Performances of Random Forest model + Naive Adaptation applied on different embedding models on Tasks 2 \& 3. Bold texts with underline style indicate the best performances.}
\label{tab:ml_task23}
\end{subtable}
\caption{Results of NLP Paradigm 1: supervised machine learning using embedding models. These are results from Random Forest models.}
\label{tab:ret_ml}
\end{table}

\begin{table}[]
\centering
\begin{tabular}{l|ccc||cccc} \hline
\multirow{2}{*}{Tasks}     & \multicolumn{3}{c||}{Datasets (\# Triples)} & \multicolumn{4}{c}{Model Peformance}   \\ \cline{2-8}
       & Training     & Validation     & Test      & Accuracy & Precision & Recall & F1     \\ \hline
Task 1 & 496,308      & 62,039         & 62,039    & 0.9565   & 0.9798    & 0.9319 & 0.9552 \\
Task 2 & 489,144      & 61,143         & 61,143    & 0.9840   & 0.9931    & 0.9749 & 0.9839 \\
Task 3 & 493,903      & 61,739         & 61,739    & 0.8723   &  0.9240   & 0.8124 & 0.8646   \\ \hline
\end{tabular}
\caption{Results of NLP paradigm 2: Fine-tuning datasets and performances of fine-tuned PubmedBERT models on three tasks}
\label{tab:ret_finetuned}
\end{table}


\begin{table}[h]
\small
\begin{subtable}[h]{\textwidth}
\centering
\begin{tabulary}{\columnwidth}{l|l|L|L|L|L|L|L}
\hline
\textbf{Model} &
  \textbf{\begin{tabular}[c]{@{}l@{}}Prompt \\ formulation \end{tabular}} &
  \textbf{\begin{tabular}[c]{@{}l@{}}Overall accuracy: \\ Mean (SD) \end{tabular}} &
  \textbf{\begin{tabular}[c]{@{}l@{}}No. unclassified \\ (\%) \end{tabular}} &
  \textbf{\begin{tabular}[c]{@{}l@{}}Precision: \\ Mean (SD) \end{tabular}} &
  \textbf{\begin{tabular}[c]{@{}l@{}}Recall: \\ Mean (SD) \end{tabular}} &
  \textbf{\begin{tabular}[c]{@{}l@{}}F1: \\ Mean (SD) \end{tabular}} &
  \textbf{Kappa} \\ \hline
\multirow{3}{*}{GPT-3.5} &
  \#1 &
  0.8040 (0.0083) &
  0 (0) &
  0.9724 (0.0007) &
  0.6518 (0.0155) &
  0.7804 (0.0114) &
  1.00 \\ \cline{2-8} & 
  \#2 & 
  0.7020 (0.0084) & 
  109 (21.8) &
  1.0000 (0.0000) & 
  0.8067 (0.0025) & 
  0.8930 (0.0015) & 
  0.98 \\ \cline{2-8} &
  \#3 &
  0.7380 (0.0045) &
  95 (19.0) &
  0.9273 (0.0135) &
  0.8485 (0.0000) &
  0.8861 (0.0062) &
  0.97 \\ \hline
\multirow{3}{*}{BioGPT} &
  \#1 &
  0.4600 (0.0255) &
  92 (18.4) &
  0.4667 (0.3613) &
  0.0412 (0.0344) &
  0.0730 (0.0580) &
  0.07 \\ \cline{2-8} & 
  \#2 & 
  0.3500 (0.0224) & 
  111 (22.2) &
  0.6333 (0.4150) & 
  0.0276 (0.0196) & 
  0.0526 (0.0369) & 
  0.05 \\ \cline{2-8} &
  \#3 &
  0.4620 (0.0356) &
  111 (22.2) &
  0.6530 (0.0892) &
  0.2872 (0.0671) &
  0.3979 (0.0814) &
  0.13 \\ \hline
  \multirow{3}{*}{GPT-4} &
  \#1 &
  0.9160 (0.0055) &
  0 (0) &
  1.0000 (0.0000) &
  0.8250 (0.0114) &
  0.9041 (0.0068) &
  0.98 \\ \cline{2-8} & 
  \#2 & 
  0.8660 (0.0152) & 
  27 (5.4) &
  0.9723 (0.0010) & 
  0.8340 (0.0183) & 
  0.8978 (0.0110) & 
  0.95 \\ \cline{2-8} &
  \textbf{\#3} &
  \textbf{0.8320 (0.0164)} &
  \textbf{55 (11.0)} &
  \textbf{1.0000 (0.0000)} &
  \textbf{0.8385 (0.0327)} &
  \textbf{0.9119 (0.0195)} &
  \textbf{0.96} \\ \hline
\end{tabulary}
\caption{Task 1 - Classification of true versus randomly generated negative triples. Relationship type: `Is\_a'.}
\end{subtable}
 \hfill
\begin{subtable}[h]{\textwidth}
\centering
\begin{tabulary}{\columnwidth}{l|l|L|L|L|L|L|L}
\hline
\textbf{Model} &
  \textbf{\begin{tabular}[c]{@{}l@{}}Prompt \\ formulation \end{tabular}} &
  \textbf{\begin{tabular}[c]{@{}l@{}}Overall accuracy: \\ Mean (SD) \end{tabular}} &
  \textbf{\begin{tabular}[c]{@{}l@{}}No. unclassified \\ (\%) \end{tabular}} &
  \textbf{\begin{tabular}[c]{@{}l@{}}Precision: \\ Mean (SD) \end{tabular}} &
  \textbf{\begin{tabular}[c]{@{}l@{}}Recall: \\ Mean (SD) \end{tabular}} &
  \textbf{\begin{tabular}[c]{@{}l@{}}F1: \\ Mean (SD) \end{tabular}} &
  \textbf{Kappa} \\ \hline
\multirow{3}{*}{GPT-3.5} &
  \#1 &
  0.6740 (0.0055) &
  0 (0) &
  0.7480 (0.0076) &
  0.6456 (0.0078) &
  0.6930 (0.0052) &
  0.97 \\ \cline{2-8} & 
  \#2 & 
  0.5920 (0.0045) & 
  97 (19.4) &
  0.7417 (0.0021) & 
  0.8446 (0.0104) & 
  0.7898 (0.0053) & 
  0.98 \\ \cline{2-8} &
  \#3 &
  0.5680 (0.0084) &
  80 (16.0) &
  0.6264 (0.0080) &
  0.8342 (0.0109) &
  0.7155 (0.0085) &
  0.98 \\ \hline
\multirow{3}{*}{BioGPT} &
  \#1 &
  0.3040 (0.0089) &
  123 (24.6)  &
  0.6667 (0.3118) &
  0.0349 (0.0186) &
  0.0656 (0.0345) &
  0.06 \\ \cline{2-8} & 
  \#2 & 
  0.4120 (0.0311) & 
  127 (25.4) &
  0.4000 (0.2937) & 
  0.0552 (0.0408) & 
  0.0968 (0.0711) & 
  0.08 \\ \cline{2-8} &
  \#3 &
  0.4180 (0.0415) &
  103 (20.6) &
  0.5877 (0.1161) &
  0.2144 (0.0650) &
  0.3111 (0.0789) &
  0.04 \\ \hline
  \multirow{3}{*}{GPT-4} &
  \#1 & 
  0.7660 (0.0134) & 
  0 (0) &
  0.7650 (0.0150) & 
  0.7680 (0.0110) & 
  0.7665 (0.0127) & 
  0.92 \\ \cline{2-8} &
  \#2 & 
  0.6880 (0.0110) & 
  43 (8.6) &
  0.7390 (0.0191) & 
  0.7753 (0.0478) & 
  0.7557 (0.0182) & 
  0.86 \\ \cline{2-8} &
  \textbf{\#3} &
  \textbf{0.8160 (0.0114)} &
  \textbf{38 (7.6)} &
  \textbf{0.8883 (0.0160)} &
  \textbf{0.8880 (0.0182)} &
  \textbf{0.8880 (0.0108)} &
  \textbf{0.94} \\ \hline
\end{tabulary}
\caption{\label{tab:res_LLM_task2} Task 2 - Classification of true versus reversed triples. Relationship type: `Is\_a'}
\end{subtable}

\hfill
\begin{subtable}[h]{\textwidth}
\centering
\begin{tabulary}{\columnwidth}{l|l|L|L|L|L|L|L}
\hline
\textbf{Model} &
  \textbf{\begin{tabular}[c]{@{}l@{}}Prompt \\ formulation \end{tabular}} &
  \textbf{\begin{tabular}[c]{@{}l@{}}Overall accuracy: \\ Mean (SD) \end{tabular}} &
  \textbf{\begin{tabular}[c]{@{}l@{}}No. unclassified \\ (\%) \end{tabular}} &
  \textbf{\begin{tabular}[c]{@{}l@{}}Precision: \\ Mean (SD) \end{tabular}} &
  \textbf{\begin{tabular}[c]{@{}l@{}}Recall: \\ Mean (SD) \end{tabular}} &
  \textbf{\begin{tabular}[c]{@{}l@{}}F1: \\ Mean (SD) \end{tabular}} &
  \textbf{Kappa} \\ \hline
\multirow{3}{*}{GPT-3.5} &
  \#1 &
  0.7180 (0.0084) &
  0 (0) &
  0.7258 (0.0128) &
  0.5773 (0.0124) &
  0.6430 (0.0110) &
  0.97 \\ \cline{2-8} & 
  \#2 & 
  0.6680 (0.0045) & 
  91 (18.2) &
  0.7838 (0.0000) & 
  0.8056 (0.0000) & 
  0.7945 (0.0000) & 
  0.99 \\ \cline{2-8} &
  \#3 &
  0.5920 (0.0110) &
  157 (31.4) &
  0.8253 (0.0089) &
  0.9393 (0.0114) &
  0.8786 (0.0062) &
  0.95 \\ \hline
\multirow{3}{*}{BioGPT} &
  \#1 &
  0.4500 (0.0520) &
  89 (17.8) &
  0.4271 (0.1920) &
  0.0664 (0.0340) &
  0.1147 (0.0576) &
  0.01 \\ \cline{2-8} & 
  \#2 & 
  0.3440 (0.0207) & 
  88 (17.6) &
  0.5833 (0.2041) & 
  0.0614 (0.0485) & 
  0.1090 (0.0827) & 
  0.03 \\ \cline{2-8} &
  \#3 &
  0.4520 (0.0319) &
  96 (19.2) &
  0.6854 (0.1036) &
  0.2989 (0.0499) &
  0.4152 (0.0650) &
  0.10 \\ \hline
  \multirow{3}{*}{GPT-4} &
  \#1 & 
  0.8740 (0.0055) & 
  0 (0) &
  0.9236 (0.0093) & 
  0.8042 (0.0114) & 
  0.8597 (0.0066) & 
  0.94 \\ \cline{2-8} &
  \#2 & 
  0.7980 (0.0045) & 
  54 (10.8) &
  0.9268 (0.0132) & 
  0.7943 (0.0128) & 
  0.8554 (0.0086) & 
  0.99 \\ \cline{2-8} &
  \textbf{\#3} &
  \textbf{0.8480 (0.0084)} &
  \textbf{24 (4.8)} &
  \textbf{0.9483 (0.0082)} &
  \textbf{0.8712 (0.0093)} &
  \textbf{0.9082 (0.0080)} &
  \textbf{0.95} \\ \hline
\end{tabulary}
\caption{Task 3 -  Classification of true versus closely-related negative triples. Relationship type: `Is\_a'}
\end{subtable}
\caption{\label{tab:res_LLM_comparision} Results of NLP paradigm 3: comparisons of effectiveness and consistency of in-context learning using LLMs for all three tasks with different prompting strategies. Note: for accuracy evaluation, the unclassified triples were included; for other metrics, those were NOT included.}
\end{table}

\begin{table}[h]
\begin{subtable}[h]{\textwidth}
\centering
\begin{tabular}{l|l|l|l|l|l}
\hline
\textbf{Model}                 & \textbf{Embeddings} & \textbf{Accuracy} & \textbf{Precision} & \textbf{Recall} & \textbf{F1 score} \\ \hline
GPT-4                          & -                   & 0.8500            & \textbf{0.9750}    & 0.8125          & 0.8864            \\ \hline
\multirow{3}{*}{Random forest} & GloVe-Chem          & \textbf{0.9600}   & 0.9607             & \textbf{0.9600} & \textbf{0.9600}   \\ \cline{2-6} 
                               & W2V-Chem            & \textbf{0.9600}   & 0.9607             & \textbf{0.9600} & \textbf{0.9600}   \\ \cline{2-6}  
                               & PubmedBERT          & 0.9400            & 0.9428             & 0.9400          & 0.9399            \\ \hline
\end{tabular}
\caption{Task 1. }
\end{subtable}

\hfill

\begin{subtable}[h]{\textwidth}
\centering
\begin{tabular}{l|l|l|l|l|l}
\hline
\textbf{Model}                 & \textbf{Embeddings} & \textbf{Accuracy} & \textbf{Precision} & \textbf{Recall} & \textbf{F1 score} \\ \hline
GPT-4                          & -                   & 0.7800            & 0.8222             & 0.7551          & 0.7872            \\ \hline
\multirow{3}{*}{Random forest} & GloVe-Chem          & 0.9300            & 0.9302             & 0.9300          & 0.9300            \\ \cline{2-6} 
                               & W2V-Chem            & 0.9100            & 0.9102             & 0.9100          & 0.9100            \\ \cline{2-6}  
                               & PubmedBERT          & \textbf{1.0000}   & \textbf{1.0000}    & \textbf{1.0000} & \textbf{1.0000}   \\ \hline
\end{tabular}
\caption{Task 2}
\end{subtable}

\hfill

\begin{subtable}[h]{\textwidth}
\centering
\begin{tabular}{l|l|l|l|l|l}
\hline
\textbf{Model}                 & \textbf{Embeddings} & \textbf{Accuracy} & \textbf{Precision} & \textbf{Recall} & \textbf{F1 score} \\ \hline
GPT-4                          & -                   & 0.8100            & 0.8723             & 0.8542          & 0.8632            \\ \hline
\multirow{3}{*}{Random forest} & GloVe-Chem          & \textbf{0.9800}   & \textbf{0.9808}    & \textbf{0.9800} & \textbf{0.9800}   \\ \cline{2-6} 
                               & W2V-Chem            & \textbf{0.9800}   & \textbf{0.9808}    & \textbf{0.9800} & \textbf{0.9800}   \\ \cline{2-6} 
                               & PubmedBERT          & 0.9500            & 0.9502             & 0.9500          & 0.9500            \\ \hline
\end{tabular}
\caption{Task 3}
\end{subtable}
\caption{Head to head comparisons of three NLP paradigms}
\label{tab:comparisons}
\end{table}

%% file: figures.tex

\begin{figure}[h]
\centering
\begin{subfigure}[t]{.7\textwidth}
        \centering
        \includegraphics[width=\textwidth]{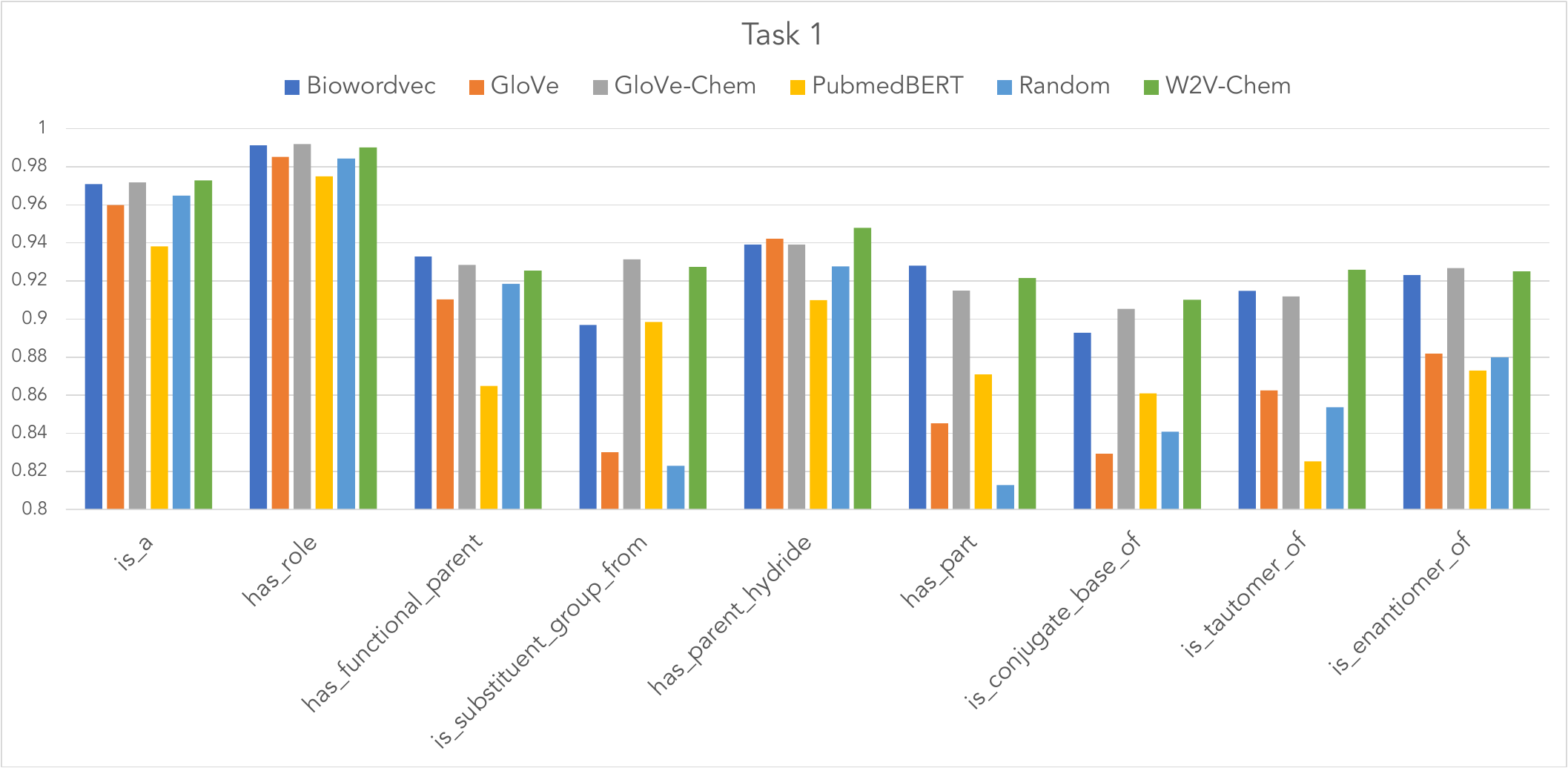}
        \caption{Task 1. }
\end{subfigure}%

\begin{subfigure}[t]{.7\textwidth}
        \centering
        \includegraphics[width=\textwidth]{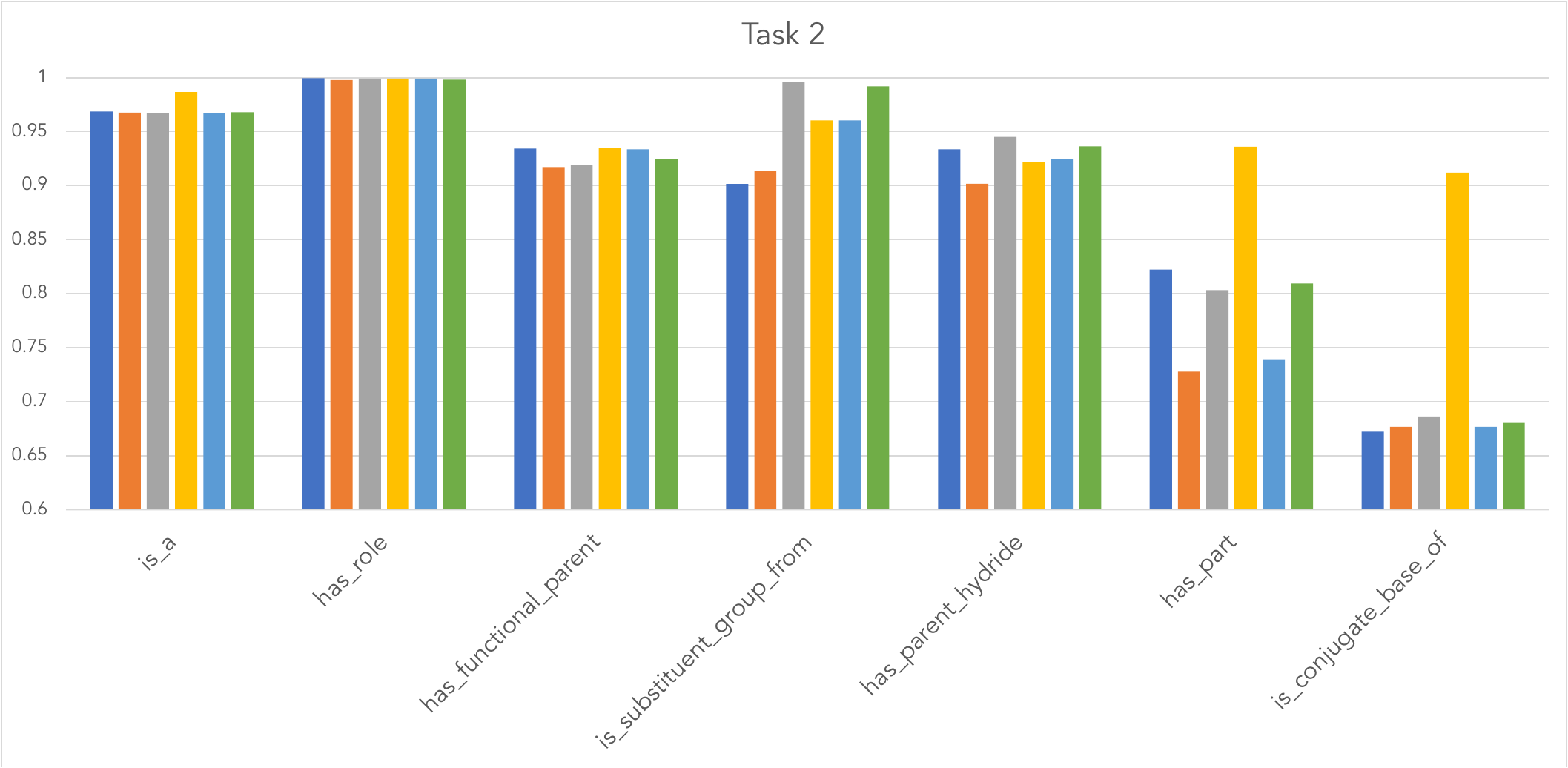}
        \caption{Task 2. }
\end{subfigure}%

\begin{subfigure}[t]{.7\textwidth}
        \centering
        \includegraphics[width=\textwidth]{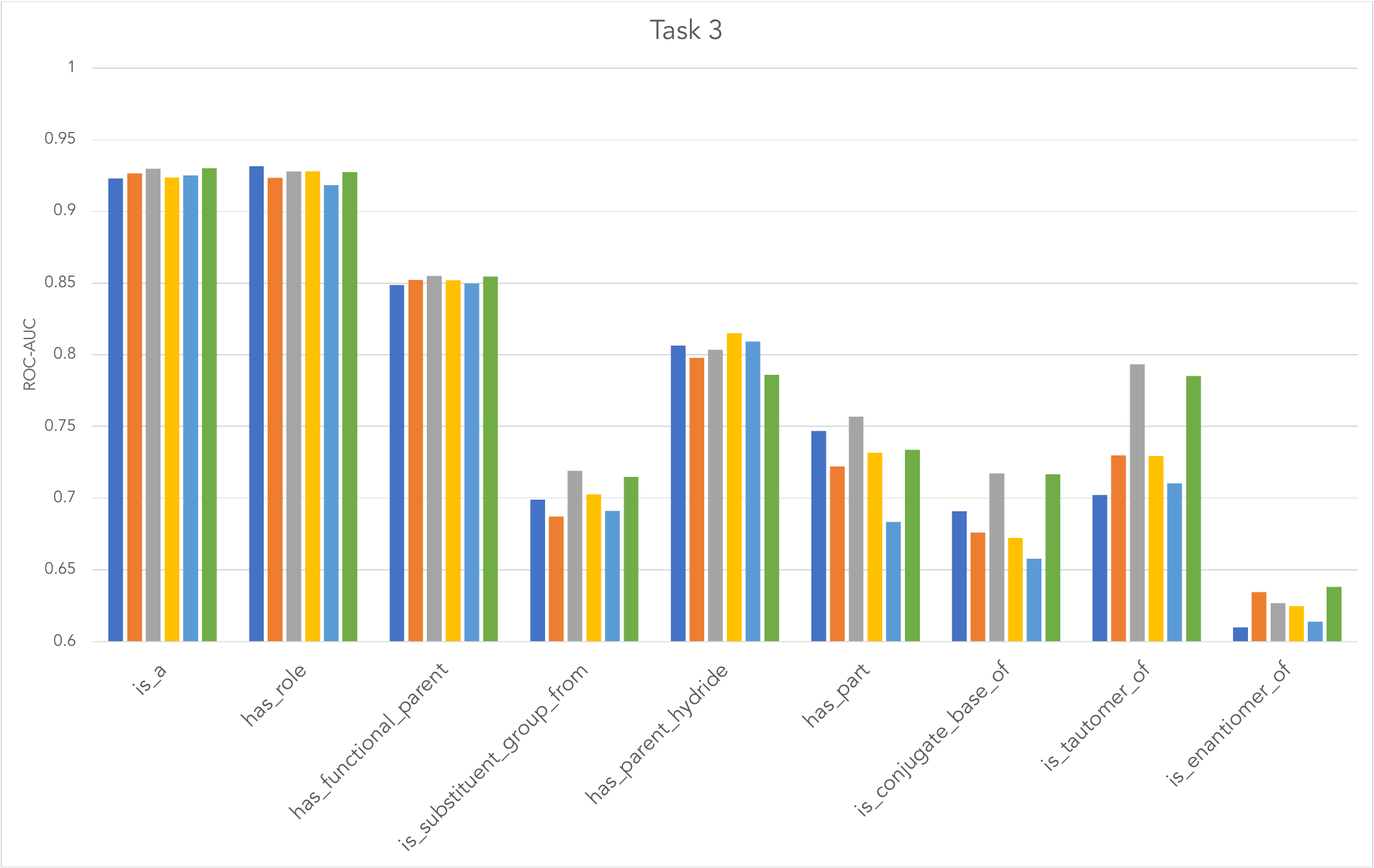}
        \caption{Task 3. }
\end{subfigure}%
\caption{Task 1-3 ROC-AUC result breakdown by relationship type. These were results from Random Forest models trained on embedding models using naive adaptation.}
\label{fig:per_relationship_results}
\end{figure}

\begin{figure}[h]
\centering
\begin{subfigure}[t]{0.45\textwidth}
        \centering
        \includegraphics[width=\textwidth]{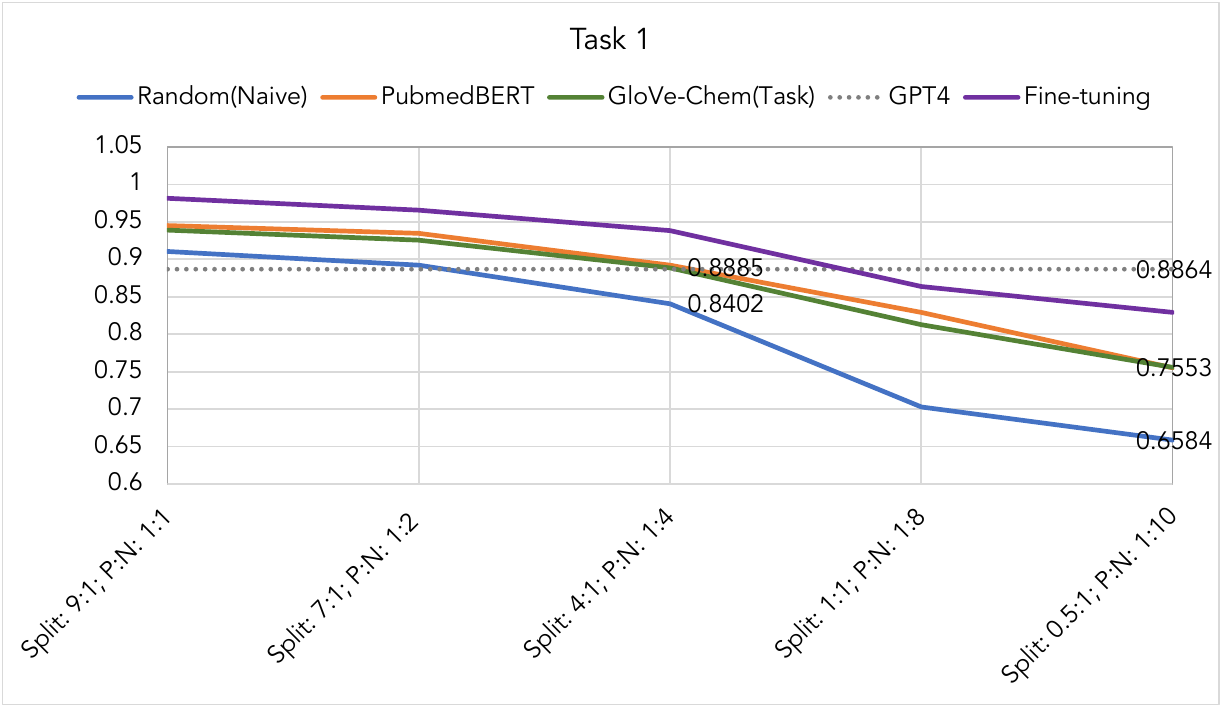}
        \caption{Task 1. Top two most consistent ML models were those using PubmedBERT  embeddingsand GloVe-Chem embeddings with task-oriented adaptation.}
\end{subfigure}%
~
\begin{subfigure}[t]{0.45\textwidth}
        \centering
        \includegraphics[width=\textwidth]{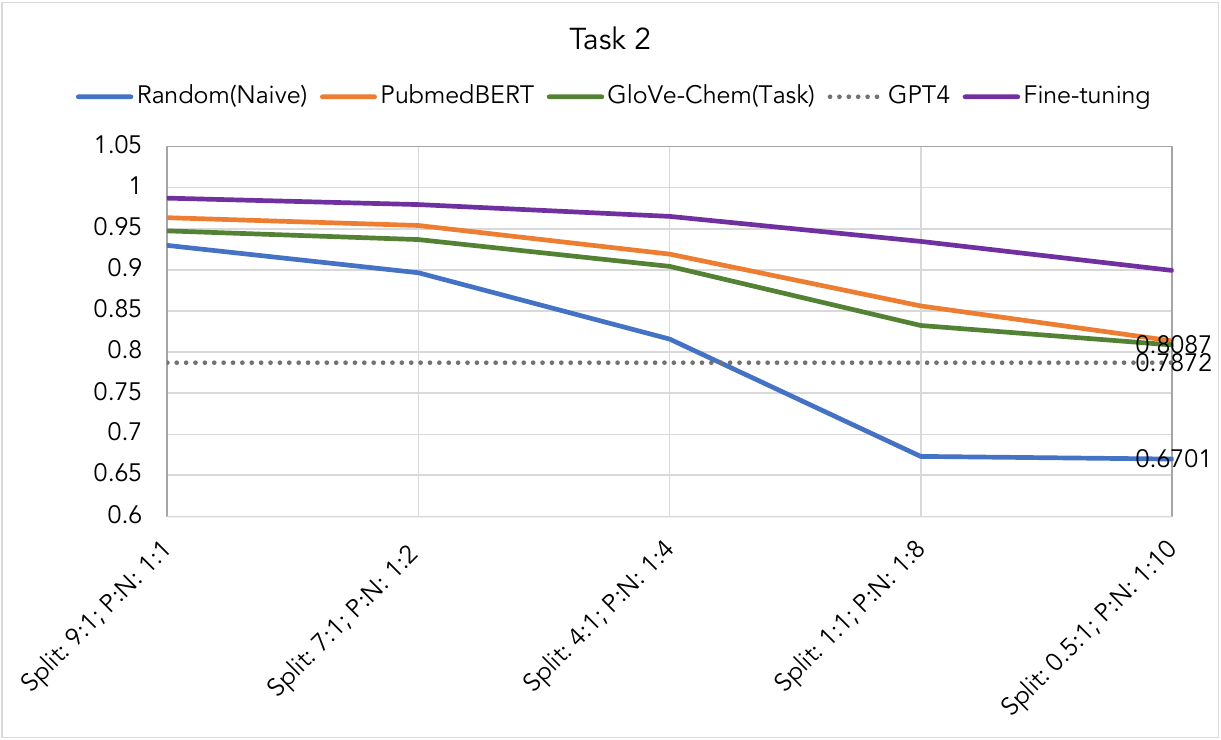}
        \caption{Task 2. Top two most consistent ML models were those using PubmedBERT embeddings and GloVe-Chem ebmeddings with task-oriented adaptation.}
\end{subfigure}%

\begin{subfigure}[t]{0.45\textwidth}
        \centering
        \includegraphics[width=\textwidth]{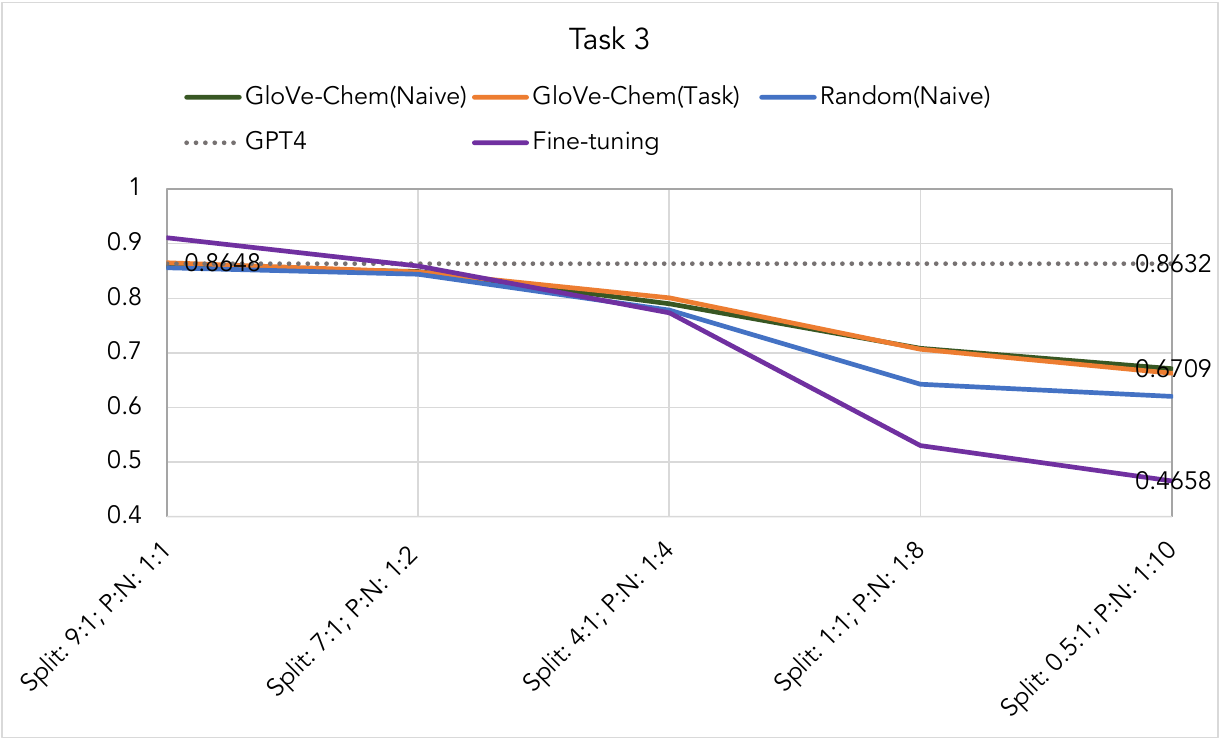}
        \caption{Task 3. Top two most consistent ML models were those using GloVe-Chem with naive-adaptation and GloVe-Chem with task-oriented adaptation}
\end{subfigure}%
\caption{F1 scores by training data volume (split) and level of imbalance (ratio of positive:negative triples) for Tasks 1-3. Graphs depict results for representative models from all three NLP paradigms. Dashed gray lines indicate GPT-4 in-context learning performances and purple lines are those of fine-tuned PubmedBERT models. Other lines are for embedding models with different adaptations, barring PubmedBERT, for which results are shown for models trained using embeddings without token selection.}
\label{fig:Imb_splits}
\end{figure}

%% file: result_ml.tex
Table~\ref{tab:ml_task1} shows the results of random forest models on task 1 - classifying positive versus random negative triples. For vanilla ML applications of embedding models (the columns titled \textbf{No adaptation}), the random embedding model performed best.  Analysis of the feature importance has led to two proposed adaptations of using embedding models (see section~\ref{sec:adaptations}): a naive adaptation and a task-oriented one. The right two columns of Table~\ref{tab:ml_task1} give performances of these adaptations. For all embedding models, both adaptations resulted in improved performances, including the model trained using random embeddings. All semantic embedding models could achieve better performances than the random model by applying adaptation approaches. W2V-Chem, a word2vect model trained from scratch, performed best with naive adaptation, while BioWordVec was the best with the task-oriented adaptation. Overall, W2V-Chem with naive adaptation was the best model for the enrichment task 1 (F1-score: 0.9690). LSTM models' performances on task 1 were listed in the Appendix Table~\ref{tab:res_task1_lstm}. Performances of LSTM models in general were on par with those of random forest models. For the simplicity of reporting and discussions, LSTM models' result were not presented and discussed in the rest of this paper.

Table~\ref{tab:ml_task23} shows the results of Random Forest with the best adaptation for task 2 and 3. In both situations, the naive adaptation gave the best results. The best performing embedding model for task 2 was PubmedBERT, while GloVe-Chem (GloVe futher trained on ChEBI related papers) was the best for task 3. Detailed results of different adaptation approaches are listed in the Appendix Table~\ref{tab:full_ml_task23}.

Comparing the best F1-scores across three tasks (Task 1: 0.9690, Task 2: 0.9822 and Task 3: 0.9125), it seemed Task 3 (in which negative triples were formed by replacing the object with similar entities) was the most challenging for ML based approaches and Task 2 (in which negative triples were formed by swapping the relationship direction) was the easiest.

Figure~\ref{fig:per_relationship_results} shows model performances by relationship type across the three tasks. In task 1, W2V-Chem performance was consistently superior to other embeddings across relationship types. GloVe-Chem and BioWordVec also performed well overall, indicating the effectiveness of task/domain related semantics in this task. For task 2, PubmedBERT embeddings were consistently better than any other model. All other models seemed to perform relatively poorly on \emph{is\_conjugate\_base\_of} ($<0.7$) and \emph{has\_part} (with two $<0.75$ and another two $<0.85$). For task 3, domain-adapted embedding models (W2V-Chem, GloVe-Chem) performed strongly for all relationship types. Three relationships appeared to be particularly challenging for all models: \emph{is\_enantiomer\_of}, \emph{is\_conjugate\_base\_of} and \emph{is\_substituent\_group\_from}.

%% file: result_llms.tex
Table~\ref{tab:res_LLM_comparision} contains results for LLM prompting experiments for classification of true versus randomly-generated negative triples (Task 1), true versus reversed triples (Task 2) or true versus closely-related negative triples (Task 3). The column \emph{No. unclassified} shows the numbers of triples for which the model either did not give a valid result (True of False) or explicitly said `I don't know' in our second prompting strategy. These triples were deemed as not accurately classified in \emph{accuracy} evaluation. However, they were excluded in \emph{precision}, \emph{recall} and \emph{F1} calculations. This was done in order to comprehensive evaluate both LLMs' general performances on all tests, and performances where a decisive answer was given.

Overall, GPT-3.5 Turbo and GPT-4 achieved competitive performances via few-shot prompting approaches. 
GPT-4 outperformed GPT-3.5 Turbo by a considerable margin in all three tasks, attaining maximal F1 scores of 0.9041, 0.8880 and 0.9082, respectively. Importantly, both models provided extremely consistent responses, with minimum Fleiss kappa scores of 0.95 and 0.86, respectively. 

Performance of BioGPT was comparatively poor, however, with accuracy and Fleiss' kappa scores consistent with random guessing. 
Modification of the base prompt to allow the models to answer `I don't know' did not appear to reliably enhance precision and F1 scores, but did generally lead to an increase in proportion of unclassified triples and consequent reduction in overall accuracy. Randomization of the ordering of positive and negative example triples appeared to be a more effective modification. In particular, GPT-4 prompted using this formulation yielded the highest F1 scores in all tasks. 


%% file: results_comparison_3paradigms.tex
Comparing results from Tables~\ref{tab:ret_ml}-\ref{tab:res_LLM_comparision}, across three enrichment tasks, supervised learning approaches and fine-tuning pretrained BERT models achieved similar performances, with the exception of task 3, where the fine-tuned PubmedBERT model performed worse. Both were superior to in-context learning of LLMs, even only considering those triples for which LLMs gave confident answers. However, these might not direct comparable because they were not assessed using the same test set. 

For a head-to-head comparison, we compared representative models from the three NLP paradigms using 100 random triples derived from the test dataset, and thus previously unseen for all models (see section~\ref{sect:preprocessing} for details). We chose GloVe-Chem and W2V-Chem as the two best models for ML based paradigm, selected the PubmedBERT embedding model to represent the fine-tuning paradigm, and GPT-4 as the representative foundation model. Table~\ref{tab:comparisons} gives the comparison results. The supervised learning paradigm achieved the best performances in 2 out of 3 tasks. Fine-tuning LLMs was the best approach in task 2. Performances attained via LLM prompting were uniformly worse than the other two paradigms: respectively 11\%, 15\% and 17\% lower in accuracy in each of the three tasks than the next best model.

\subsubsection{Effects of imbalanced data and variations in train-test splits on supervised learning and fine-tuning}
For the two NLP paradigms that might be affected, we further examined the effects of training models using smaller, more imbalanced datasets. Using a random 10\% of the full dataset, we trained and evaluated ML models in five different scenarios, from abundant/balanced to successively less training data/more imbalanced, as follows. 
\begin{itemize}
    \item Task 1. A constant test size of 6,204, with five different training set sizes of 55,835, 43,427, 24,815, 6,204, 3,102.
    \item Task 2. A constant test size of 6,114, with five different training set sizes of 55,029, 42,800, 24,457, 6,114, 3,057.
    \item Task 3. A constant test size of 6,174, with five different training set sizes of 55,564, 42,800, 24,457, 6,174, 3,087.
\end{itemize}

Figure~\ref{fig:Imb_splits} shows the changing patterns of F1-scores of three representative models in all three tasks (full data was depicted in Appendix Figure~\ref{fig:all_data_variants}). For each task, we picked three models; models trained using random vectors (as a reference) and two most consistently performing models. Unsurprising, in all tasks, performances decreased steadily as less training data was available and greater imbalance was introduced. PubmedBERT and GloVe-Chem were the most consistent models, i.e., less prone to sub-optimal training data. Fine-tuned models outperformed all ML based approaches in the first two tasks. However, the fine-tuned approach suffered much more significantly for task 3, performing worse even than ML with random embeddings.

We also plotted GPT-4's performances on the figures. Essentially, GPT-4 would be a better tool to use in scenarios where those solid lines are below the dashed line, i.e., ML based or fine-tuning approaches won't achieve any better performances than GPT-4. For task 1, GPT-4 outperformed both ML-based and fine-tuned approaches in the two most extreme scenarios, i.e.,  (Split: 1:1; P:N: 1:8) and (Split: 0.5:1; P:N: 1:10). For task 3, GPT-4 was superior in all but the first setting (Split: 9:1; P:N: 1:1). For task 2, however, GPT-4 never surpassed ML-based or fine-tuned approaches.

In all tasks, random forest models trained on random embeddings started as a strong baseline, achieving comparable performances to best models. However, when data availability decreased, their performance dropped much more significantly versus other models.


%% file: appendix.tex
\setcounter{table}{0}
\renewcommand{\thetable}{A\arabic{table}}

\setcounter{figure}{0}
\renewcommand{\thefigure}{A\arabic{figure}}

\begin{table}[h]
\begin{tabular}{@{} p{2cm}|p{7.5cm}|p{7cm} @{}}
\hline
\textbf{Sub-ontology} & \textbf{Definition} &
\textbf{Example} \\ 
\hline
Chemical entities & Classifies molecular entities (or parts of entities) according to their composition and structure & Hydrocarbons, carboxylic acids, tertiary amines   \\ \hline
Role entities & Classifies entities on the basis of their role within: (i) a chemical context; (ii) a biological context; or (iii) intended use by humans & (i) Ligand, inhibitor, surfactant; (ii) antibiotic, antiviral agent, coenzyme, hormone; (iii) pesticide, antirheumatic drug, fuel
\\ \hline
Subatomic particles   & Classifies sub-atomic particle entities & Electron, photon, nucleon           \\ \hline
\end{tabular}
\caption{\label{tab:summary_subontologies}Included ChEBI sub-ontologies.}
\label{tab.chebi}
\end{table}

\begin{table}[h]
\begin{tabular}{@{} p{4cm}|p{6.5cm}|p{6cm} @{}}
\hline
\textbf{Relationship} & \textbf{Description} &
\textbf{Example} \\ 
\hline
Is a & Defines the relationship between more specific and more general concepts & Tetrabutylammonium fluoride is a fluoride salt                   
\\ \hline
Has part & Defines the relationship between part and whole & Cobalt dichloride has part cobalt(2+)
\\ \hline
Is conjugate base of & Defines the relationship between acids and their conjugate bases & Mannarate(1-) is conjugate base of mannaric acid
\\ \hline
Is tautomer of & Defines the cyclic relationship used to show the interrelationship between two tautomers & 2-mercaptosuccinate is tautomer of 3-carboxy-2-sulfidopropanoate
\\ \hline
Is enantiomer of & Defines the cyclic relationship used in instances when two entities are non-superimposable mirror images of each other & Dexverapamil hydrochloride is enantiomer of (S)-verapamil hydrochloride 
\\ \hline
Has functional parent & Defines the relationship between two molecular entities or classes of entities, of which one possesses one or more characteristic groups from which the other can be derived by functional modification & Vecuronium bromide has functional parent 5alpha-androstane
\\ \hline
Has parent hydride & Defines the relationship between an entity and its parent hydride & Serpentine has parent hydride 18-oxayohimban
\\ \hline
Is substituent group from & Defines the relationship between a substituent group or atom and its parent molecular entity, from which it is formed by loss of one or more protons or simple groups such as hydroxyl groups & N(2)-L-glutamino(1-) group is substituent group from L-glutaminate
\\ \hline
Has role & Defines the relationship between a molecular entity and the particular behaviour it may exhibit (either by nature or by human application) & Ammonium chloride has role ferroptosis inhibitor
\\ \hline
\end{tabular}
\caption{\label{tab:summary_relationships}Included ChEBI relationship types.}
\label{tab.rels}
\end{table}

\begin{table}
\begin{center}
\begin{tabular}{l|r}
\hline
\textbf{Relationship type} & \textbf{Number of triples} \\ \hline
is a                       & 230,241                  \\ \hline
has role                   & 42,095                   \\ \hline
has functional parent      & 18,204                   \\ \hline
is conjugate base of       & 8,247                    \\ \hline
is conjugate acid of       & 8,247                    \\ \hline
has part                   & 3,911                    \\ \hline
is enantiomer of           & 2,674                    \\ \hline
is tautomer of             & 1,804                    \\ \hline
has parent hydride         & 1,736                    \\ \hline
is substituent group from  & 1,279                    \\ \hline \hline
\textbf{Total \#triples}  & \textbf{318,438}                   \\ \hline
\end{tabular}
\end{center}
\captionsetup{justification=centering}
\caption{Numbers of triples in ChEBI.}
\label{tab:summary_relationships_freqs}
\end{table}

\begin{table}[h!]
\centering
\begin{tabular}{l|r|c|r}
\hline
\textbf{Embedding model}     & \textbf{Vocubulary size} & \textbf{Dimensions} & \textbf{OOV (\%)} \\ \hline
GloVe                  & 2,196,017       & 300                 & 41,887 (87.81)     \\ \hline
W2V-Chem   & 151,563        & 300                 & 33,952 (71.18)     \\ \hline
GloVe-Chem      & 2,276,964       & 300                 & 30,632 (64.22)     \\ \hline
Biowordvec              & 2,347,646       & 200                 & 22,797 (47.79)     \\ \hline
PubmedBERT             & 28,895       & 768            
  & -                  \\ \hline
\end{tabular}
\caption{\label{tab:embeddings_basic}Embedding model size and out of vocabulary (OOV) statistics.}
\end{table}

\begin{figure}[h]
\centering
\includegraphics[width=.95\linewidth]{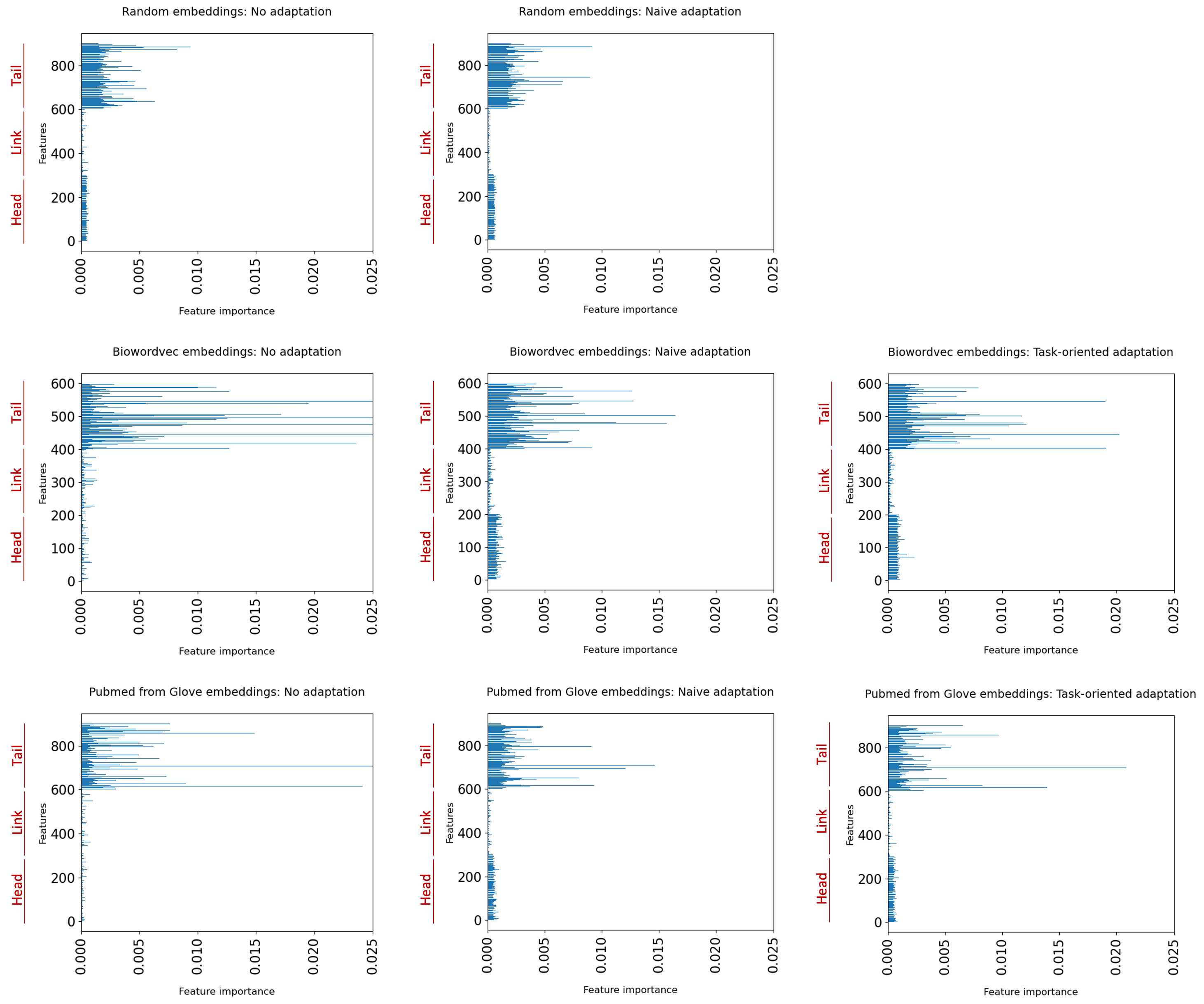}
\caption{Feature importance patterns of random forest models trained on different embedding models and using different adaptation methods. Figures in the first row are those generated of random embeddings. The second row lists those of Biowordvec models while the last row shows GloVe-Chem models (GloVe embedding further trained on ChEBI related PubMed articles). The firs column lists those models without adaptation, while the middle shows those of naive adaptation and the rightmost one lists those using taks-orientied adaptation.}
\label{fig:feature_importance_task1_allTokens}
\end{figure}

\begin{table}[H]
\begin{tabulary}{\columnwidth}{|l|L|}
\hline
\textbf{}     & \textbf{50 most frequently-occurring tokens}     \\ \hline
\textbf{Head} & {[}'2', '3', '4', '1', '5', '6', 'yl', 'n', 'd', 'methyl', 'hydroxymethyl', '6r', '2s', '2r', '3r', 'beta', '5s', '3s', '8', '4s','oxan', '4r', 'oxy', '10', 'oxo', '12', 'acid', '5r', '0', 'hydroxy', 'alpha', 'dihydroxy', '7', '9', '14', 'acetamido', 'l', '15', '13', 'amino', 'trihydroxy', 'methoxy', 'hydroxypropan', '6s', 'o', 'trien', 'phenyl', '11', 'acetamide', 'dihydro'{]}                                                                                                                   \\ \hline
\textbf{Tail} & {[}'acid', '1', 'metabolite', '3', 'd', '2', 'compound', '4', 'beta', 'amino', 'n', 'alpha', 'fatty', 'organic', 'l', 'lactam', 'azamacrocycle', 'peptide', 'agent', 'aromatic', 'ester', 'hydroxy', 'acyl', 'derivative', 'anion', '5', 'steroid', 'inhibitor', 'sn', 'entity', 'ether', 'coa', 'chain', 'molecular', 'oligosaccharide', 'monocarboxylic', 'galactosyl', '6', 'acetyl', 'plant', 'alcohol', 'c', 'sulfonamide', 'glycero', 'phosphate', 'human', 'amide', 'carbohydrate', 'ec', 'alkaloid'{]}\\ \hline
\end{tabulary}
\caption{\label{tab:res_top_frequent_tokens}The top 50 common tokens in head and tail entities from all positive triples.}
\end{table}

\begin{table}[H]
\centering
\begin{tabulary}{\columnwidth}{|l|l|L|L|L|}
\hline
\textbf{Model type} & \textbf{Embeddings} & \textbf{Precision} & \textbf{Recall} & \textbf{F1} \\ \hline
\multirow{5}{*}{LSTM} & Random              & 0.9517 & 0.9516 & 0.9516 \\ \cline{2-5} 
                      & GloVe               & 0.9559 & 0.9559 & 0.9559 \\ \cline{2-5} 
                      & W2v-Chem & 0.9497 & 0.9496 & 0.9496 \\ \cline{2-5} 
                      & GloVe-Chem   & 0.9538 & 0.9538 & 0.9538 \\ \cline{2-5} 
                      & \textbf{BioWordVec} & \textbf{0.9636} & \textbf{0.9636} & \textbf{0.9636} \\ \hline
\end{tabulary}
\caption{\label{tab:res_task1_lstm}Task 1 results of LSTM models trained on different embedding models.}
\end{table}

\begin{table}[]
\scriptsize
\begin{tabular}{l|ccc|ccc||ccc|ccc}  \hline
 \multirow{3}{*}{}   & \multicolumn{6}{c||}{Task 2}                                                                                                                                                            & \multicolumn{6}{c}{Task 3}                                                                                                                                                            \\ \hline
           & \multicolumn{3}{c|}{Naive adaptation}                                                      & \multicolumn{3}{c||}{Task-oriented adaptation}                                              & \multicolumn{3}{c|}{Naive adaptation}                                                      & \multicolumn{3}{c}{Task-oriented adaptation}                                              \\ \hline
           & \multicolumn{1}{l}{Precision} & \multicolumn{1}{l}{Recall} & \multicolumn{1}{l|}{F1-Score} & \multicolumn{1}{l}{Precision} & \multicolumn{1}{l}{Recall} & \multicolumn{1}{l||}{F1-Score} & \multicolumn{1}{l}{Precision} & \multicolumn{1}{l}{Recall} & \multicolumn{1}{l|}{F1-Score} & \multicolumn{1}{l}{Precision} & \multicolumn{1}{l}{Recall} & \multicolumn{1}{l}{F1-Score} \\ \hline
Random     & 0.9581                        & 0.9581                     & 0.9581                       & -                             & -                          & -                            & 0.9042                        & 0.9042                     & 0.9042                       & -                             & -                          & -                            \\
GloVe      & 0.9573                        & 0.9573                     & 0.9573                       & \textbf{0.9639}               & \textbf{0.9639}            & \textbf{0.9639}              & 0.9073                        & 0.9073                     & 0.9073                       & \textbf{0.9067}               & \textbf{0.9067}            & \textbf{0.9067}              \\ 
W2V-Chem   & 0.9596                        & 0.9596                     & 0.9596                       & 0.9507                        & 0.9507                     & 0.9507                       & 0.9122                        & 0.9122                     & 0.9122                       & 0.8786                        & 0.8773                     & 0.8779                       \\
GloVe-Chem & 0.9586                        & 0.9586                     & 0.9586                       & 0.9725                        & 0.9725                     & 0.9725                       & {\underline{\textbf{0.9126}}}         & {\underline{\textbf{0.9125}}}      & {\underline{\textbf{0.9125}}}        & 0.9051                        & 0.9051                     & 0.9051                       \\
BioWordVec & 0.9605                        & 0.9605                     & 0.9605                       & 0.9595                        & 0.9595                     & 0.9595                       & 0.9062                        & 0.9061                     & 0.9061                       & 0.894                         & 0.8937                     & 0.8938                       \\ 
PubmedBERT & {\underline{\textbf{0.9822}}}         & {\underline{\textbf{0.9822}}}      & {\underline{\textbf{0.9822}}}        & -                             & -                          & -                            & 0.906                         & 0.906                      & 0.9060                       & -                             & -                          & -                           \\ \hline
\end{tabular}
\caption{Random forest models' performances on tasks 2 \& 3 using different adaptation methods.}
\label{tab:full_ml_task23}
\end{table}

\begin{figure}[h]
\centering
\begin{subfigure}[t]{0.5\textwidth}
        \centering
        \includegraphics[width=\textwidth]{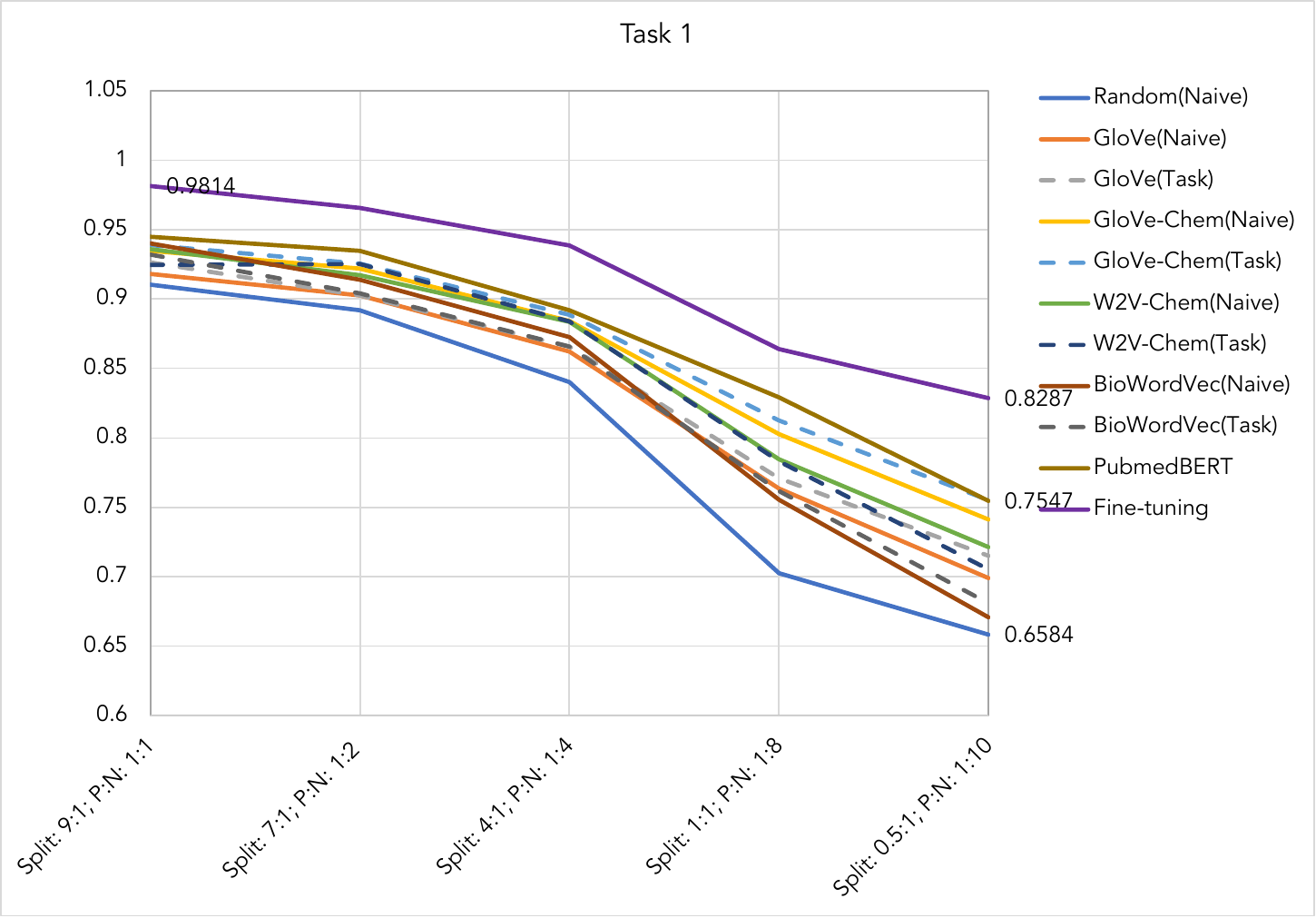}
        \caption{Task 1}
\end{subfigure}%
~
\begin{subfigure}[t]{0.5\textwidth}
        \centering
        \includegraphics[width=\textwidth]{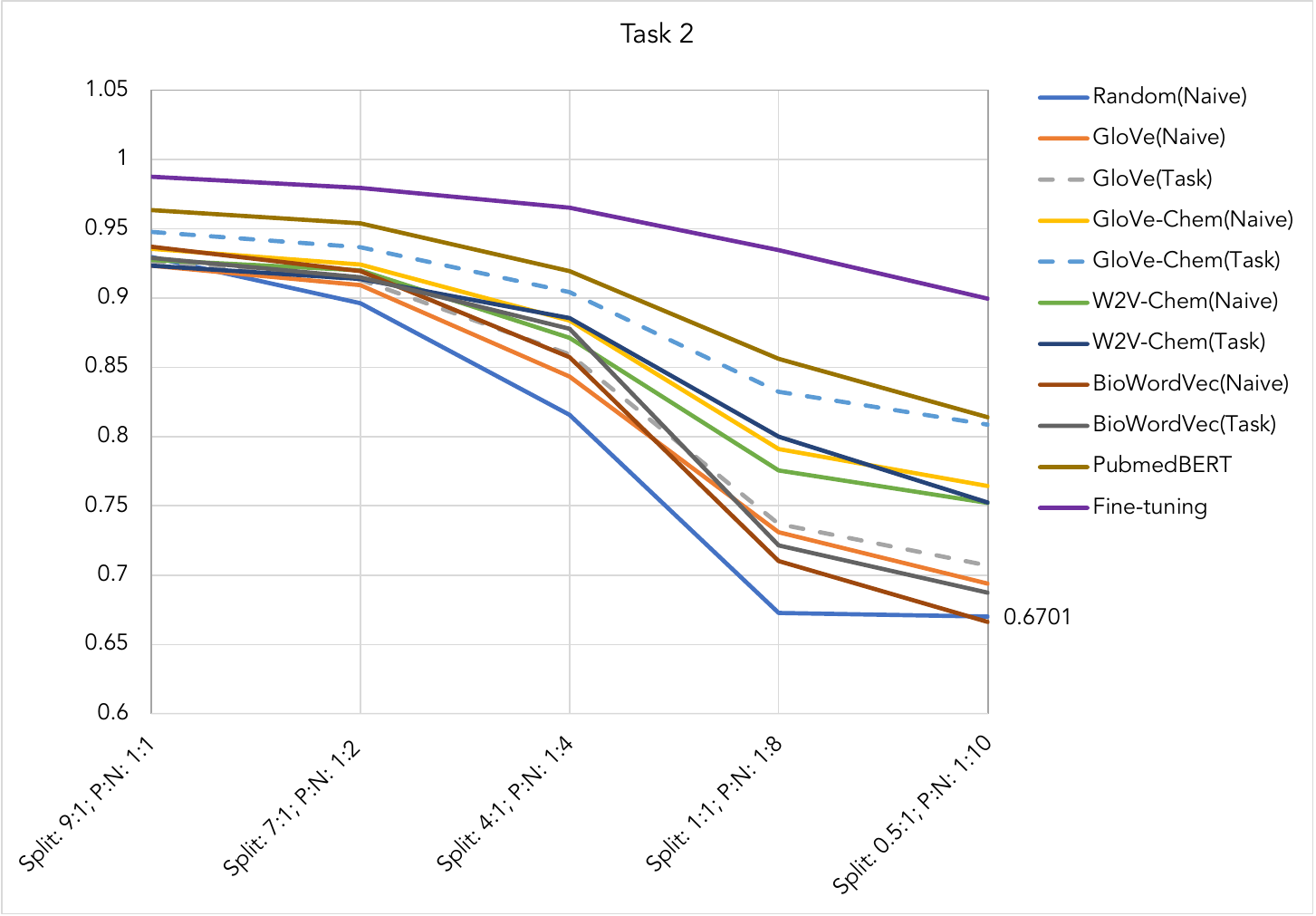}
        \caption{Task 2}
\end{subfigure}%

\begin{subfigure}[t]{0.5\textwidth}
        \centering
        \includegraphics[width=\textwidth]{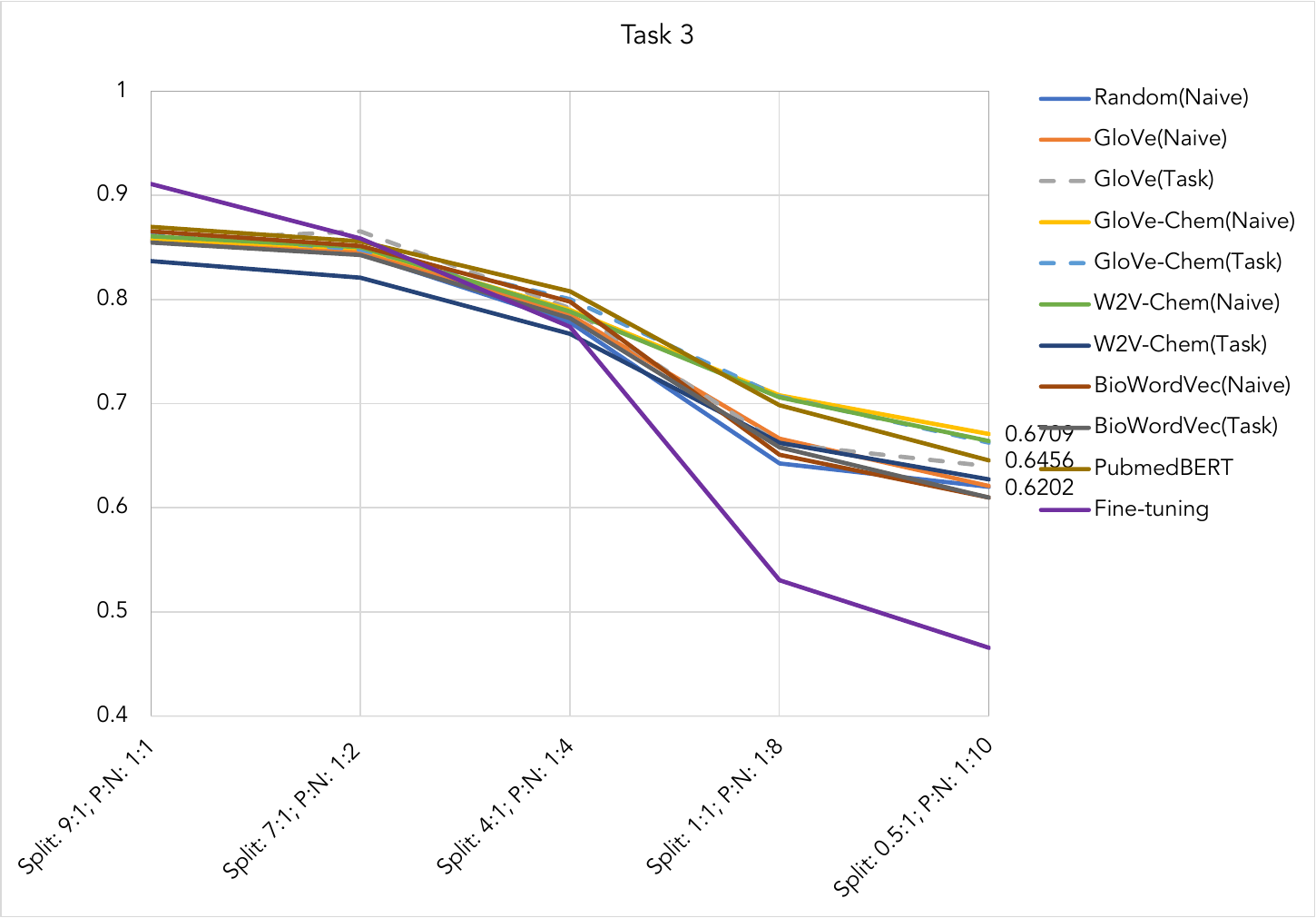}
        \caption{Task 3}
\end{subfigure}%
\caption{F1 scores by training data volume (split) and level of imbalance (ratio of positive:negative triples) for Tasks 1-3. Graphs depict results for embedding models with naive (token length-based) adaptation, barring PubmedBERT, for which results are shown for models trained using embeddings without stop-word selection.}
\label{fig:all_data_variants}
\end{figure}